\documentclass[runningheads]{llncs}

\usepackage{eccv}

\usepackage{xspace}
\newcommand{\CohD}{\ensuremath{\mathrm{Coh}_{\mathrm{D}}}\xspace}
\newcommand{\CohH}{\ensuremath{\mathrm{Coh}_{\mathrm{H}}}\xspace}
\newcommand{\CohV}{\ensuremath{\mathrm{Coh}_{\mathrm{V}}}\xspace}
\newcommand{\VDEI}{\ensuremath{\mathrm{VD\text{-}EI}}\xspace}
\newcommand{\TUNNEL}{\texttt{SpatialTunnel}\xspace}

\usepackage{eccvabbrv}

\usepackage{graphicx}
\usepackage{booktabs}
\usepackage{wrapfig}

\usepackage[accsupp]{axessibility}  

\usepackage{multirow}
\usepackage{xcolor}

\usepackage[pagebackref,breaklinks,colorlinks,citecolor=eccvblue]{hyperref}

\usepackage{orcidlink}

\begin{document}

\title{Why Far Looks Up: Probing Spatial Representation in Vision-Language Models} 

\titlerunning{Why Far Looks Up}

\author{Cheolhong Min\inst{1} \and
Jaeyun Jung\inst{1} \and
Daeun Lee\inst{1} \and
Hyeonseong Jeon\inst{1} \and \\
Yu Su\inst{2} \and
Jonathan Tremblay\inst{3} \and
Chan Hee Song\inst{3, \dagger, \ddagger} \and
Jaesik Park\inst{1, \dagger} 
}

\institute{Seoul National University \and 
The Ohio State University \and 
\noindent NVIDIA \\
\email{lusong@nvidia.com, \{cheolhong.min, jaesik.park\}@snu.ac.kr}
}

\def\thefootnote{$^\dagger$}\footnotetext{Co-corresponding author}
\def\thefootnote{$^\ddagger$}\footnotetext{Project Lead}


\authorrunning{C. Min et al.}


\maketitle

\begin{abstract}
Vision-language models (VLMs) achieve strong performance on spatial reasoning benchmarks, yet it remains unclear whether this reflects structured 3D understanding or reliance on statistical shortcuts in natural images.
We introduce a representation-level analysis framework that constructs minimal contrastive pairs to measure how spatial axes are organized and disentangled within VLM embeddings.
Our analysis across multiple model families reveals a consistent \emph{vertical-distance entanglement}: models conflate vertical image position with distance, mirroring the perspective bias of natural photographs. 
This bias produces a significant accuracy gap between perspective-consistent and counter-heuristic examples, and intensifies under data scaling even as overall benchmark accuracy improves.
We further show that models with similar benchmark scores can exhibit different internal representations, and that these differences predict accuracy and robustness across diverse spatial reasoning benchmarks.
To isolate this bias from evaluation-set skew, we introduce \TUNNEL, a synthetic benchmark designed to expose spatial shortcut biases by removing common correlations present in natural images.
Experiments suggest that the entanglement is model-intrinsic, and that models with well-separated spatial axes exhibit greater robustness, indicating that well-structured spatial representations lead to more reliable spatial reasoning across diverse benchmarks.
Code and benchmark are available on the 
\href{https://cheolhong0916.github.io/whyfarlooksup.github.io/}{project website}.

\end{abstract}



\section{Introduction}
\label{sec:intro}

\begin{figure}
    \centering
    \includegraphics[trim={0mm 1mm 0mm 0mm}, clip, width=1.00\linewidth]{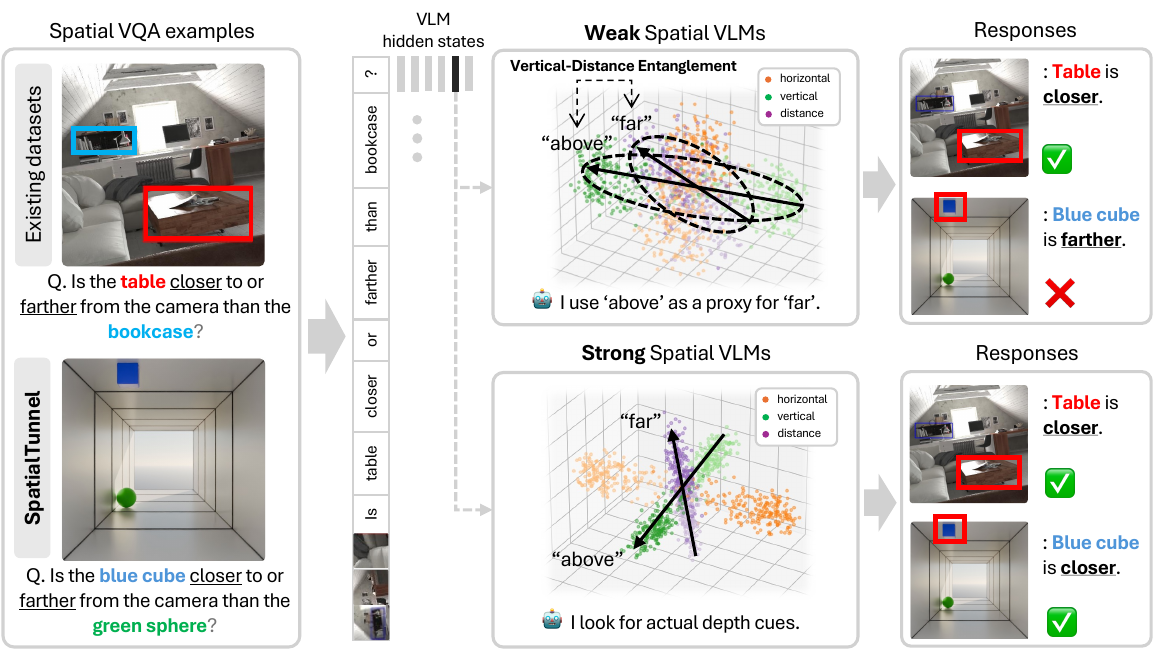}
    \caption{Many VLMs answer spatial questions via a perspective-driven shortcut, \eg., objects located higher in the image are further away in 3D. 
    By confusing 2D vertical position with 3D distance, models fail systematically on counter examples.
    Our \TUNNEL benchmark and contrastive probing expose this vertical-distance entanglement.
    In contrast, strong spatial VLMs show disentangled axes and consistent correctness across both real and synthetic settings.}
    \label{fig:placeholder}
    \vspace{-1em}
\end{figure}

Spatial reasoning is a core capability for Vision-Language Models (VLMs), particularly as these systems are increasingly deployed in robotics~\cite{team2025gemini, kim24openvla, nvidia2025gr00tn1openfoundation, intelligence2025pi05visionlanguageactionmodelopenworld}, embodied agents~\cite{llm-planner, singh2023progprompt, ahn2022can}, and multimodal assistants~\cite{anthropic2025claude, singh2025openaigpt5card, comanici2025gemini25pushingfrontier} that observe and interact with physical environments. 
Although modern VLMs are primarily trained on 2D image--text pairs~\cite{bai2025qwen3, Qwen2.5-VL, LLaVA-1.5, deitke2025molmo}, they achieve strong performance on spatial reasoning benchmarks~\cite{du2024embspatial, fu2024blink, tong2024cambrian}, and recent work continues to improve these results through scaling and spatial training data~\cite{tan2026robobrain25depthsight, zhou2025roborefer, cheng2024spatialrgpt, song2025robospatial, chen2026spacetools}. 
These advances suggest that current models possess meaningful spatial understanding. 
However, it remains unclear whether strong benchmark accuracy reflects robust spatial reasoning or the exploitation of statistical regularities in natural images.

Many spatial relations can be partially inferred from correlations that arise naturally in photographic data rather than from explicit reasoning about 3D spatial structure. 
For example, perspective in everyday photographs introduces a consistent relationship between vertical image position and depth: objects appearing higher in the image are often farther from the camera, as in Figure~\ref{fig:placeholder}. 
Such correlations allow models to rely on shortcuts that substitute vertical cues for depth reasoning, achieving high benchmark accuracy while internally conflating distinct spatial dimensions.

This limitation highlights a broader challenge in evaluating spatial understanding in VLMs.
Behavioral benchmarks measure whether a model produces correct answers, but they provide limited insight into \emph{how} those answers are obtained.
Two models may achieve similar performance while relying on different internal mechanisms: one encoding spatial relations in a structured, separable manner, and another depending on correlated cues present in natural imagery which become brittle under distribution shift.
Distinguishing these possibilities requires examining how spatial information is represented inside the model, rather than relying on output-level performance.

Recent work has revealed persistent spatial reasoning failures through controlled benchmarks~\cite{zhang2025do, kamath2023whatsup, zhang2025mllmsstrugglespatialunderstanding} and has begun probing internal model behavior such as attention dynamics~\cite{chen2025why}.
However, these efforts primarily assess individual task performance or local mechanisms, leaving the global geometric organization of spatial relations in representation space largely unexplored.

We address this gap from two complementary angles.
First, we analyze how spatial relations along three core 3D axes -- horizontal (left / right), vertical (above / below), and depth (close / far) -- are organized within VLM internal embeddings, using controlled contrastive examples that vary only the spatial relation between objects while holding 
confounds such as object identity fixed.
Second, we introduce \texttt{SpatialTunnel}, a synthetic benchmark designed to remove perspective-driven biases in spatial evaluation. Its tunnel geometry decouples vertical image position from depth, enabling balanced assessment beyond the correlations present in natural image benchmarks.

Across multiple VLM families, our experiments reveal that horizontal relations form stable, opposing directions in representation space, whereas vertical and depth relations are frequently entangled, suggesting reliance on perspective-driven cues.
Moreover, models with more structured spatial representations perform better across diverse spatial reasoning benchmarks, including EmbSpatial-Bench~\cite{du2024embspatial}, CV-Bench~\cite{tong2024cambrian}, and BLINK~\cite{fu2024blink}.
Evaluations on \texttt{SpatialTunnel} further expose biases hidden under standard benchmark settings, and models with more structured representations exhibit greater robustness when these correlations are removed.
Together, these results suggest that benchmark accuracy alone may overestimate the spatial reasoning capabilities of current VLMs.
Our contributions are threefold:
\begin{itemize}
    \item \textbf{Representation-level analysis of spatial reasoning.} 
    We introduce a framework for analyzing how spatial relations are organized within VLM embeddings, diagnosing whether models encode structured spatial reasoning or rely on shortcut cues.

    \item \textbf{Spatial representations predict robustness.} 
    We show that models with similar benchmark performance can exhibit markedly different internal spatial representations, and that models with more structured spatial representations exhibit greater robustness and generalization.

    \item \textbf{A bias-controlled synthetic benchmark for spatial reasoning.} 
    We construct a synthetic dataset that decouples vertical image position from depth, revealing shortcut biases hidden under standard benchmark settings.
\end{itemize}

\section{Related Work}
\label{sec:related}

\paragraph{\textbf{Spatial Understanding Datasets and Benchmarks.}}
Recent benchmarks have revealed persistent weaknesses in VLM spatial reasoning despite strong semantic performance.
Controlled evaluations such as What's Up~\cite{kamath2023whatsup} and COMFORT~\cite{zhang2025do} show that models frequently fail on basic positional distinctions and frame-of-reference consistency.
To probe deeper spatial competence, subsequent work has expanded along several axes: egocentric and cross-video reasoning~\cite{du2024embspatial, tong2024cambrian}, 6DoF diagnostic tasks~\cite{wang2025spatial457}, and multi-step spatial referring~\cite{zhou2025roborefer}.
In parallel, simulation-based datasets~\cite{ray2025sat, song2025robospatial, team2025gemini} provide large-scale supervision for physical dynamics, yet spatial performance often plateaus with data scaling~\cite{zhang2025mllmsstrugglespatialunderstanding}.
While these efforts effectively measure \emph{whether} models succeed or fail, they do not examine \emph{what cues} models rely on internally—in particular, none isolate the entanglement between vertical image position and perceived depth that arises from perspective projection.
Our work targets this gap by constructing controlled synthetic environments and contrastive splits that systematically expose this bias.

\paragraph{\textbf{Probing Internal Representations of Vision-Language Models.}}
Recent work has moved beyond behavioral evaluation to examine the internal states of VLMs.
Linear probing studies show that vision encoders inherently represent monocular depth cues~\cite{danier2025depthcues} and bind geometric coordinates to object activations in early layers~\cite{kang2026linearprobing}, while unified extraction frameworks~\cite{sheta2025behavioral} facilitate systematic comparison across model families.
On the mechanistic side, ADAPTVIS~\cite{chen2025why} analyzes attention dynamics during spatial reasoning, and Spatial Forcing~\cite{li2025spatialforcing} explicitly aligns intermediate layers with 3D structure.
However, these approaches primarily detect the \emph{presence} of individual spatial primitives or adjust local attention behavior; they do not examine how different spatial dimensions are \emph{jointly organized}—in particular, whether depth and vertical cues occupy separable or entangled directions in representation space.
We address this gap through controlled contrastive analysis of internal embeddings, directly measuring the geometric relationship between spatial axes to reveal entanglement that isolated probing cannot detect.

\section{Perspective Projection Bias in Spatial Understanding}
\label{sec:finding}



Vision-language models are increasingly expected to reason about 3D spatial relationships from a single RGB image, \eg, answering questions such as ``Is the chair closer to the camera than the table?''
However, monocular images provide only a 2D projection of the 3D scene, requiring models to infer spatial structure from indirect visual cues.
A central question is whether current VLMs genuinely learn such 3D reasoning, or instead rely on the visual cues that happen to correlate with depth in image space.

In this section, we analyze how VLMs perform spatial reasoning across multiple model families and benchmarks.
Our analysis reveals a systematic bias arising from perspective projection: models frequently use an object's vertical position in the image as a proxy for its distance from the camera.
We term this phenomenon \textbf{vertical-distance entanglement}, where image-plane vertical position becomes conflated with depth.
Across multiple models and benchmarks, we show that this bias consistently emerges and leads to systematic errors in spatial reasoning.

\subsection{What is Vertical-Distance Entanglement?}
\label{sec:definition}

\begin{wrapfigure}[11]{R}{0.45\linewidth}
\centering
\vspace{-2.8em}
\includegraphics[width=\linewidth]{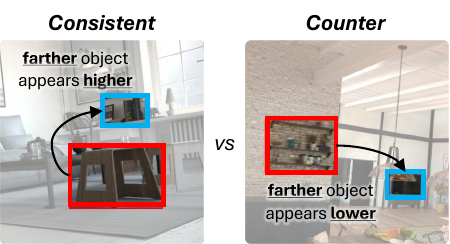}
\caption{\textbf{\textit{Consistent} vs.\ \textit{counter} examples.} \textbf{\textit{Consistent}}: Farther object appears higher in the image; \textbf{\textit{Counter}}: Farther object appears lower.}
\label{fig:cc_examples}
\end{wrapfigure}

\paragraph{\textbf{Perspective projection and vertical position.}}
From the observer’s viewpoint, objects farther away on a common ground surface appear higher in the image.
This phenomenon gives rise to the classical elevation cue: for objects lying on the ground plane, those nearer to the horizon line are perceived as being farther from the observer~\cite{danier2025depthcues} (see Appendix~\hyperref[anchor:appA]{\textcolor{red}{A}} for details).

\paragraph{\textbf{Entanglement as a shortcut.}}
We hypothesize that VLMs exploit this correlation as a shortcut: when asked about relative depth, they partially rely on the vertical positions of objects rather than reasoning about 3D structure.
We refer to this phenomenon as \textbf{vertical-distance entanglement}, indicating the tendency of a model to treat \textit{above} $\approx$ \textit{far} and \textit{below} $\approx$ \textit{close} when answering depth-related questions.

\paragraph{\textbf{Consistent and counter-heuristic examples.}}

To systematically analyze this entanglement, we categorize depth-related samples into two groups, \emph{consistent} and \emph{counter} (Figure~\ref{fig:cc_examples}). The classification is based on whether the ground-truth spatial relationship aligns with the vertical-position heuristic.

We implement this by comparing the vertical center coordinates of the two queried objects in pixel space: if the farther object has a smaller $y$-coordinate (\ie, higher in the image), the example is consistent; otherwise, it is counter.
If a model exhibits no entanglement, its accuracy should be comparable on both groups.
Conversely, a systematic accuracy gap between the two groups would constitute evidence that the model relies on the vertical-position shortcut.

\subsection{Experimental Setup}
\label{sec:exp_setup}


\paragraph{\textbf{Models.}}
We evaluate three VLM families spanning different architectures: Molmo-7B-O-0924~\cite{deitke2025molmo}, NVILA-Lite-2B~\cite{liu2025nvila}, and Qwen2.5-VL-3B-Instruct~\cite{Qwen2.5-VL}.
To analyze how spatial fine-tuning affects entanglement, we train variants of each model at multiple data scales (80k, 400k, 800k, and 2M samples); base models refer to the original pretrained weights without additional fine-tuning.
We also include RoboRefer-2B-SFT~\cite{zhou2025roborefer}, which shares the NVILA-Lite-2B base but is trained on more than 20M samples including RGB and RGB-D images, and Qwen3-VL-235B-A22B-Instruct~\cite{bai2025qwen3} as a large-scale reference.

\paragraph{\textbf{Training data.}}
Recent work has attributed VLMs' limited spatial understanding to a lack of spatial reasoning data during training, motivating several spatial-focused datasets~\cite{song2025robospatial, chen2024spatialvlm, ray2025sat, zhang2025flatland, zhou2025roborefer, deshpande2025graspmolmo}.
To study the effect of data scaling within and across model families, we uniformly mix five existing spatial understanding datasets (\ie, SAT~\cite{ray2025sat}, RoboSpatial~\cite{song2025robospatial}, SPAR-7M~\cite{zhang2025flatland}, RefSpatial~\cite{zhou2025roborefer}, PRISM~\cite{deshpande2025graspmolmo}) and subsample at four target scales (80k, 400k, 800k, and 2M) for supervised fine-tuning (see Appendix~\ref{app:training_data} and~\ref{app:data_mix} for details).


\subsection{Evidence from Existing Benchmarks}
\label{sec:real_evidence}

We first examine whether vertical-distance entanglement is observable on established spatial reasoning benchmarks that use real-world images: EmbSpatial-Bench~\cite{du2024embspatial} and the 3D-spatial split of CV-Bench~\cite{tong2024cambrian}.

\begin{table}[t!]
\centering
\caption{\textbf{Distribution of consistent, counter, and ambiguous examples.} Existing spatial benchmarks are skewed toward consistent examples, mirroring the natural statistics of perspective projection in real-world images.}
\label{tab:distribution}
\small
\begin{tabular}{lccc}
\toprule
\textbf{Type} & \textbf{EmbSpatial-Bench} & \textbf{CV-Bench-3D} & \textbf{Definition} \\
\midrule
Consistent & 976 (80.9\%) & 363 (60.5\%) & GT aligns with heuristic \\
Counter    & 129 (10.7\%) & 65 (10.8\%)  & GT contradicts heuristic \\
Ambiguous  & 101 (8.4\%)  & 172 (28.7\%) & $\Delta y < 5\%$ of image height \\
\bottomrule
\end{tabular}
\end{table}

\begin{table}[t!]
\centering
\caption{\textbf{Accuracy on consistent vs.\ counter examples across models and benchmarks.} All models exhibit a substantial accuracy gap, with consistent examples outperforming counter examples. Results are reported on depth-related questions from EmbSpatial-Bench and CV-Bench-3D. Indented rows denote fine-tuned variants at the given spatial data scale.}
\label{tab:real_accuracy}
\small
\begin{tabular}{lcccc}
\toprule
 & \multicolumn{2}{c}{\textbf{EmbSpatial-Bench}} & \multicolumn{2}{c}{\textbf{CV-Bench-3D}} \\
\cmidrule(lr){2-3} \cmidrule(lr){4-5}
\textbf{Model} & Consistent & Counter & Consistent & Counter \\
\midrule
Molmo-7B-O-0924 \cite{deitke2025molmo} & 63.5 & 34.9 \textcolor{red}{(-28.6)} & 93.1 & 75.4 \textcolor{red}{(-17.7)} \\
~~~~+ 80k       & 60.6 & 29.5 \textcolor{red}{(-31.1)} & 80.2 & 56.9 \textcolor{red}{(-23.3)} \\
~~~~+ 400k      & 62.7 & 27.1 \textcolor{red}{(-35.6)} & 89.5 & 56.9 \textcolor{red}{(-32.6)} \\
~~~~+ 800k      & 65.2 & 34.1 \textcolor{red}{(-31.1)} & 88.7 & 70.8 \textcolor{red}{(-17.9)} \\
~~~~+ 2M        & 65.3 & 39.5 \textcolor{red}{(-25.8)} & 90.6 & 72.3 \textcolor{red}{(-18.3)} \\
\midrule
NVILA-Lite-2B \cite{liu2025nvila} & 49.0 & 27.1 \textcolor{red}{(-21.9)} & 74.4 & 40.0 \textcolor{red}{(-34.4)} \\
~~~~+ 80k       & 57.7 & 15.5 \textcolor{red}{(-42.2)} & 71.6 & 50.8 \textcolor{red}{(-20.8)} \\
~~~~+ 400k      & 61.1 & 34.1 \textcolor{red}{(-27.0)} & 81.3 & 58.5 \textcolor{red}{(-22.8)} \\
~~~~+ 800k      & 63.2 & 38.8 \textcolor{red}{(-24.4)} & 84.6 & 67.7 \textcolor{red}{(-16.9)} \\
~~~~+ 2M        & 60.7 & 41.1 \textcolor{red}{(-19.6)} & 97.2 & 93.8 \textcolor{red}{(-3.4)} \\
\midrule
RoboRefer-2B-SFT~\cite{zhou2025roborefer} & 87.0 & 59.7 \textcolor{red}{(-27.3)} & 98.9 & 95.4 \textcolor{red}{(-3.5)} \\
\midrule
Qwen2.5-VL-3B-Instruct \cite{Qwen2.5-VL} & 54.7 & 32.6 \textcolor{red}{(-22.1)} & 75.5 & 55.4 \textcolor{red}{(-20.1)} \\
~~~~+ 80k       & 50.6 & 30.2 \textcolor{red}{(-20.4)} & 69.7 & 60.0 \textcolor{red}{(-9.7)} \\
~~~~+ 400k      & 52.6 & 27.1 \textcolor{red}{(-25.5)} & 65.8 & 58.5 \textcolor{red}{(-7.3)} \\
~~~~+ 800k      & 55.8 & 26.4 \textcolor{red}{(-29.4)} & 61.2 & 58.5 \textcolor{red}{(-2.7)} \\
~~~~+ 2M        & 60.9 & 24.0 \textcolor{red}{(-36.9)} & 62.0 & 53.8 \textcolor{red}{(-8.2)} \\
\midrule
Qwen3-VL-235B-A22B-Instruct \cite{bai2025qwen3} & 73.3 & 41.7 \textcolor{red}{(-31.6)} & 98.1 & 90.8 \textcolor{red}{(-7.3)} \\
\bottomrule
\end{tabular}
\end{table}

\paragraph{\textbf{Data distribution is skewed toward consistent examples.}}
We classify all depth-related questions in both benchmarks into consistent, counter, and ambiguous categories following the criteria defined in \Cref{sec:definition}.
As shown in \Cref{tab:distribution}, consistent examples account for 80.9\% of EmbSpatial-Bench and 60.5\% of CV-Bench-3D, while counter examples constitute only about 10\% in each.
This heavy skew reflects the natural statistics of real-world photographs: in most everyday scenes, farther objects do appear higher in the image.

\paragraph{\textbf{Models systematically fail on counter examples.}}
We evaluate a range of VLMs spanning different architectures and scales on the two benchmarks, reporting accuracy separately for consistent and counter subsets (\Cref{tab:real_accuracy}).
Across \emph{all} models and \emph{all} training scales, accuracy on consistent examples significantly exceeds that on counter examples.
For instance, Qwen2.5-VL fine-tuned on 2M samples achieves 60.9\% on the consistent split of EmbSpatial-Bench but only 24\% on counter examples, yielding a 36.9 percentage-point gap.
This pattern holds regardless of model family (Molmo, NVILA, or Qwen2.5-VL), model size, or the amount of spatial fine-tuning data, suggesting that vertical-distance entanglement is a widespread phenomenon rather than an artifact of any single architecture, training recipe, or data scale.

\section{Behavioral Analysis with a Synthetic Dataset}
\label{sec:spatialtunnel}

The accuracy gap in Section~\ref{sec:finding} indicates that VLMs systematically fail on counter examples in real-world datasets.
However, real photographs conflate multiple depth cues (\eg, vertical position, apparent size, and occlusion), making it difficult to isolate the contribution of any single cue.
To enable controlled interventions, we introduce \TUNNEL, a synthetic dataset that decouples an object’s vertical image-plane position from its 3D depth by design, allowing the two factors to be manipulated independently.

\subsection{\TUNNEL Benchmark}
\label{sec:spatialtunnel_dataset}

To evaluate spatial relations in a controlled manner, we require an environment with two key properties:
(i) objects can be positioned arbitrarily, enabling queries over any spatial relation (\eg, left/right, above/below, near/far); and
(ii) an object’s vertical position can be adjusted independently of its depth, allowing us to construct image groups that differ only in vertical placement while preserving depth ordering.

\begin{figure}[t]
    \centering
    \includegraphics[width=1.0\linewidth]{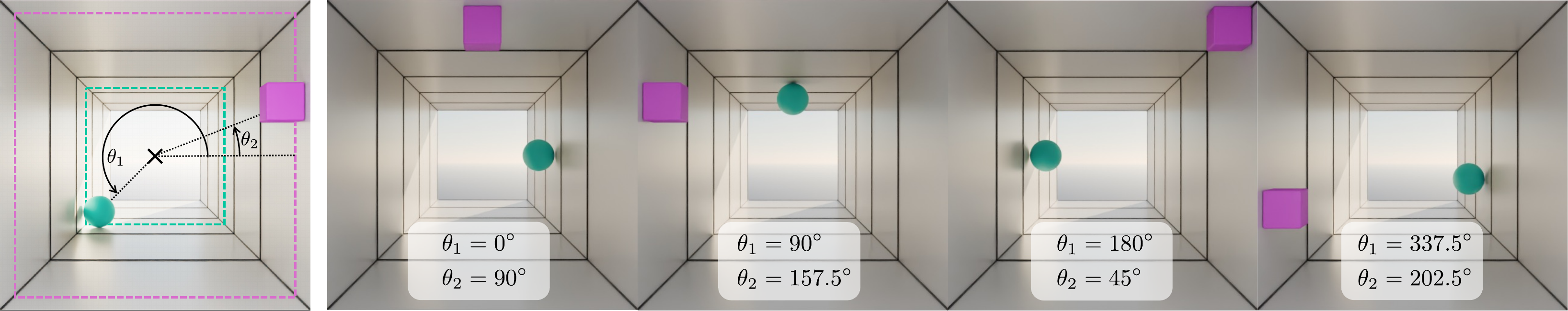}
    \caption{\TUNNEL holds the two objects at fixed depths while sweeping their angular positions around the tunnel cross-section, so that 2D image-plane layout varies independently of depth ordering.
    }
    \label{fig:tunnel_grid}
    \vspace{-1em}
\end{figure}

To satisfy these requirements, we build a tunnel-shaped synthetic scene in Blender~\cite{blender} (Figure~\ref{fig:tunnel_grid}).
Each scene consists of a single-point-perspective corridor whose walls, ceiling, and floor are symmetric about the camera's optical axis, where objects are placed anywhere on the interior tunnel surfaces.
Because objects near the top and bottom of the image can be equidistant from the camera, the common heuristic ``higher in the image $\Rightarrow$ farther'' no longer holds.
We parameterize each object by its depth $z$ and an angular position $\theta$ on the tunnel cross-section. Holding $z$ fixed while varying $\theta$ moves the object up/down and left/right in the image without changing its depth ordering, enabling matched counterfactual pairs that flip vertical arrangement while preserving depth.

We construct a synthetic benchmark suite, \TUNNEL, that enables controlled spatial interventions in a single-point-perspective corridor. 
Specifically, we place two objects at predetermined depths and sweep each object along the tunnel cross-section, discretizing the interior into 16 angular positions (see Figure~\ref{fig:tunnel_grid}). 
This yields a $16 \times 16$ Cartesian grid over $(\theta_1,\theta_2)$, enabling heatmap-style diagnostics of model behavior across configurations (see Figure~\ref{fig:heatmap_1616}). 
To increase visual diversity and improve robustness, we randomize object appearance (color, size, and shape) and scene lighting across renders.
Additional synthetic variants for other spatial cues (\eg, object size) and auxiliary analyses are provided in the Appendix~\ref{app:objsize}.

    

\subsection{Experimental Setup on \TUNNEL}
\label{sec:spatialtunnel_setup}


Given a rendered RGB image containing two objects, the model is asked a binary depth-comparison question.
In our setup, an object is always placed farther from the camera than the other, and the VLM is asked to answer the questions like ``\textit{Is \{\textit{obj$_1$}\} closer to / farther from the camera than \{\textit{obj$_2$}\}?}''
Following prior work~\cite{hu2023prompting, wang2025logical, zhang2025do}, we define a local probability by extracting the logits for \texttt{Yes} and \texttt{No} at the first generated token. We then compute the predicted probability as
\[
p=\sigma\!\bigl(\ell_{\texttt{Yes}}-\ell_{\texttt{No}}\bigr).
\]
The correctness score for a single query is defined as $v=p$ if the ground-truth answer is \texttt{Yes}, and $v=1-p$ if it is \texttt{No}.
We report the following metrics for all VLMs described in Section~\ref{sec:exp_setup}. Following the definition in Section~\ref{sec:definition}, samples are partitioned into \emph{consistent} and \emph{counter} subsets.
We report four metrics:
(1)~{Mean accuracy} ($v$), the mean correctness score across all images and questions;
(2)~{Consistent accuracy} ($v_\text{cons}$), the mean correctness score on \emph{consistent} examples;
(3)~{Counter accuracy} ($v_\text{ctr}$), the mean correctness score on \emph{counter} examples; and
(4)~{Accuracy gap} ($\Delta = v_\text{cons} - v_\text{ctr}$), the accuracy difference between the two subsets, quantifying the vertical-distance entanglement. A model with no directional bias would yield $\Delta \approx 0$.



\subsection{Results on \TUNNEL: Vertical-Distance Entanglement}
\label{sec:results_spatialtunnel}

\begin{figure}[t]
    \begin{subfigure}[t]{0.499\linewidth}
        \centering
        \includegraphics[width=1.0\linewidth]{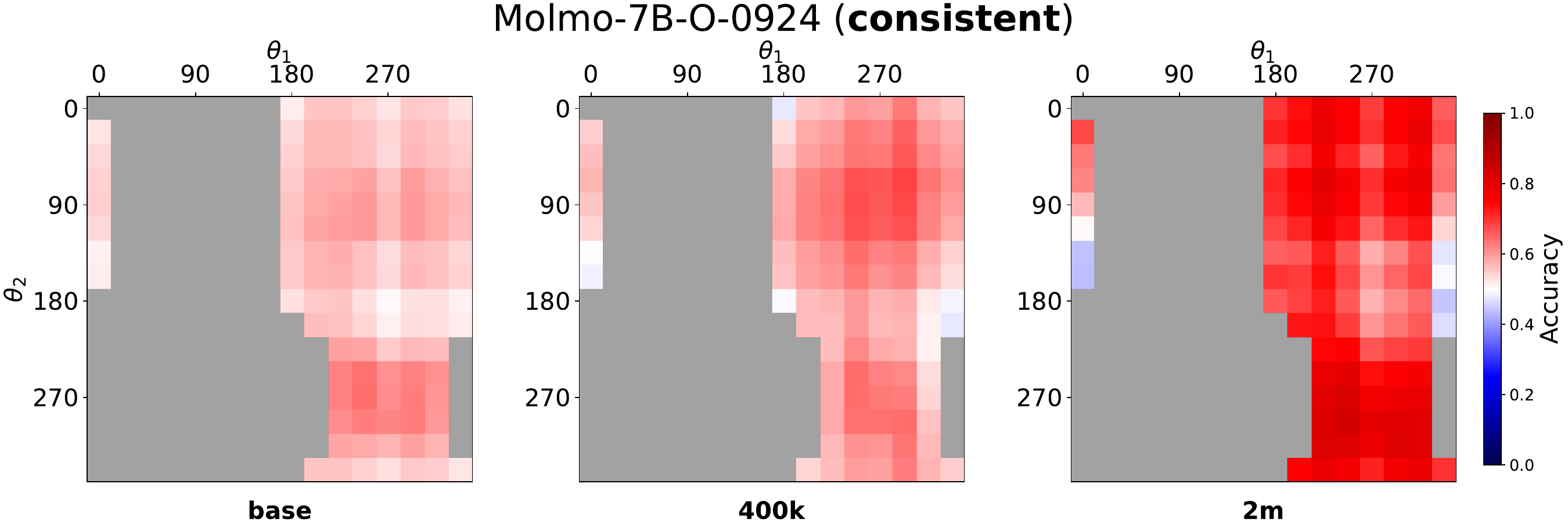}
        \caption{Results on \emph{consistent} samples}
        \label{fig:heatmap_consistent}
    \end{subfigure}
    \begin{subfigure}[t]{0.499\linewidth}
        \centering
        \includegraphics[width=\linewidth]{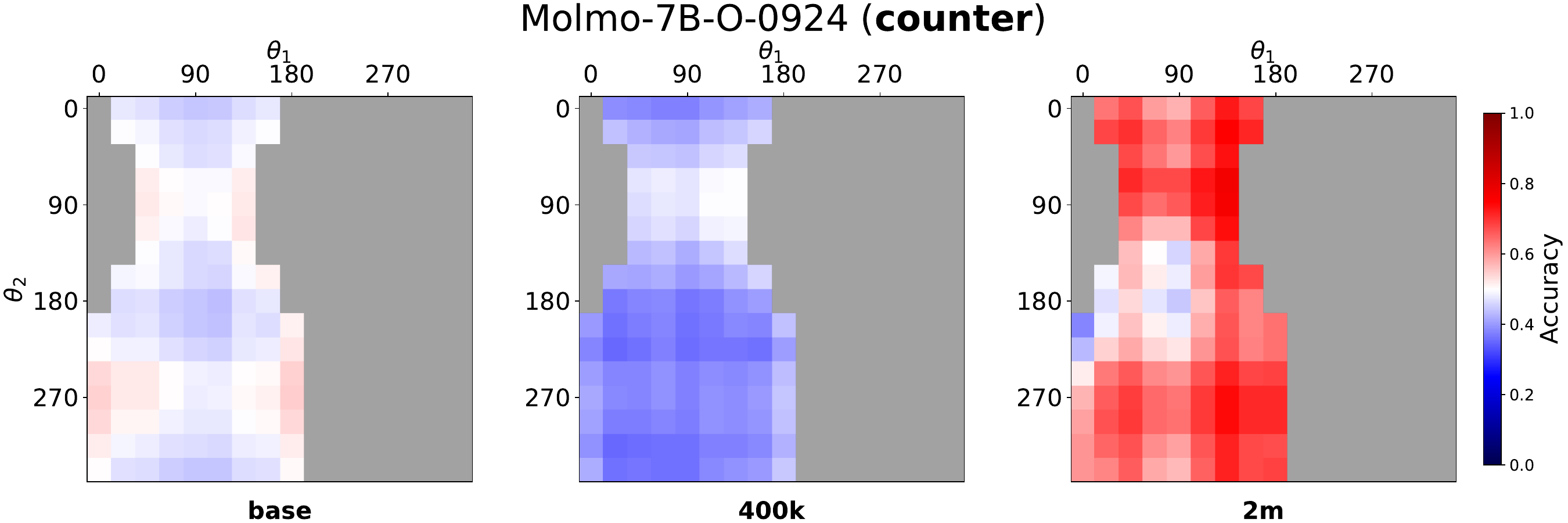}
        \caption{Results on \emph{counter} samples.} 
        \label{fig:heatmap_counter}
    \end{subfigure}
    \caption{Mean accuracy heatmaps on \TUNNEL for Molmo-7B. Each cell indexes a joint angular configuration $(\theta_1,\theta_2)$ of the two objects (red = higher accuracy; blue = lower). Gray indicates configurations outside the subset. From base $\rightarrow$ 400k $\rightarrow$ 2M training samples, accuracy on (a) perspective-consistent cells improves steadily. In contrast, (b) counter cells remain substantially harder, with the largest drop at 400k and a partial recovery at 2M.}
    \label{fig:heatmap_1616}
    \vspace{-2em}
\end{figure}

\begin{wrapfigure}[19]{r}{0.50\linewidth}
\vspace{-3.4em}
\centering
\captionof{table}{%
    \textbf{Consistent vs.\ Counter accuracy on \TUNNEL.}
    $v$: mean correctness score; $v_\text{cons}$ and $v_\text{ctr}$: scores on consistent and counter subsets; $\Delta = v_\text{cons} - v_\text{ctr}$.
}
\label{tab:spatialtunnel}
\scriptsize
\begin{tabular*}{\linewidth}{@{\extracolsep{\fill}}lcccc@{}}
\toprule
\textbf{Model} & $\mathbf{v}$ & 
$\mathbf{v_\text{cons}}$ & $\mathbf{v_\text{ctr}}$ & $\mathbf{\Delta}$ \\
\midrule
Molmo-7B-O-0924    & 0.528 & 0.565 & 0.487 & +0.078 \\
~~~~+ 80k   & 0.496 & 0.507 & 0.486 & +0.021 \\
~~~~+ 400k  & 0.501 & 0.593 & 0.409 & +0.184 \\
~~~~+ 800k  & 0.531 & 0.628 & 0.430 & +0.198 \\
~~~~+ 2M    & 0.666 & 0.703 & 0.630 & +0.073 \\
\midrule
NVILA-Lite-2B & 0.488 & 0.504 & 0.471 & +0.033 \\
~~~~+ 80k  & 0.499 & 0.562 & 0.438 & +0.124 \\
~~~~+ 400k  & 0.669 & 0.804 & 0.538 & +0.267 \\
~~~~+ 800k  & 0.646 & 0.728 & 0.571 & +0.157 \\
~~~~+ 2M    & 0.812 & 0.875 & 0.749 & +0.127 \\
\midrule
RoboRefer-2B-SFT & 0.793 & 0.816 & 0.770 & +0.046 \\
\midrule
Qwen2.5-VL-3B & 0.570 & 0.776 & 0.360 & +0.416 \\
~~~~+ 80k   & 0.512 & 0.585 & 0.440 & +0.145 \\
~~~~+ 400k  & 0.503 & 0.588 & 0.418 & +0.171 \\
~~~~+ 800k  & 0.499 & 0.600 & 0.398 & +0.202 \\
~~~~+ 2M    & 0.500 & 0.648 & 0.353 & +0.295 \\
\midrule
Qwen3-VL-235B & 0.908 & 0.948 & 0.880 & +0.068 \\
\bottomrule
\end{tabular*}
\end{wrapfigure}

Consistent with Section \ref{sec:real_evidence}, we observe that the vertical-distance entanglement is universal. Across all base and fine-tuned models, accuracy is consistently higher on the consistent subset than on the counter subset, yielding a positive accuracy gap $\Delta$.
Table~\ref{tab:spatialtunnel} and Figure~\ref{fig:heatmap_1616} summarize model behavior on \TUNNEL.

For example, base Qwen2.5-VL-3B achieves $v_\text{cons} = 0.776$ but only $v_\text{ctr} = 0.360$, indicating strong reliance on the vertical-position shortcut. 
While base NVILA-Lite-2B produces a narrower gap, its sub-0.5 overall accuracy suggests near-random performance rather than meaningful depth understanding.
Figure~\ref{fig:heatmap_1616} visualizes positional bias at the cell level for Molmo-7B variants. 
If predictions were insensitive to 2D placement, accuracy would be approximately uniform across the grid. Instead, most models show pronounced contrast between consistent and counter regions.
The results suggest that large-scale spatial training reduces this reliance. RoboRefer~\cite{zhou2025roborefer}, trained on more than 20M QA pairs, achieves the smallest gap ($\Delta=+0.046$) among models performing above chance. Qwen3-VL-235B attains the highest mean accuracy ($v = 0.908$) with a similarly small gap ($\Delta = +0.068$), indicating that very large-scale pretraining can substantially alleviate this bias even without targeted spatial fine-tuning.

\label{sec:representation}

\begin{figure}
    \centering
    \includegraphics[width=1.0\linewidth]{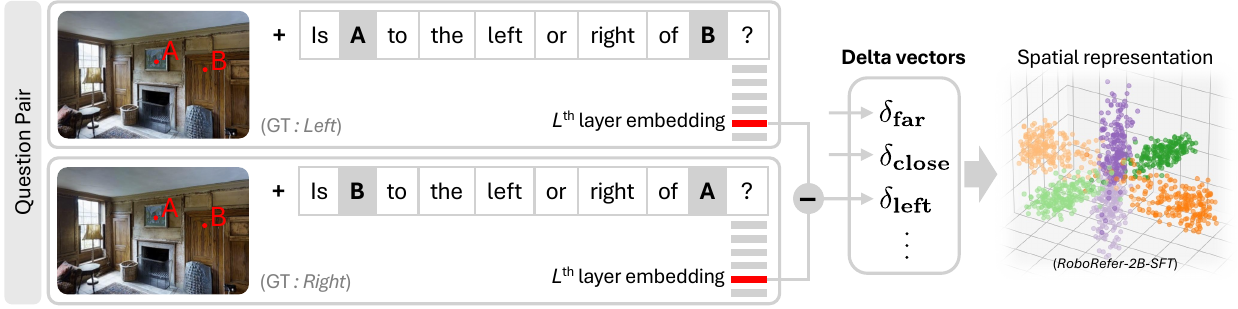}
    \caption{\textbf{Contrastive probing for representation-level spatial analysis.}
Given a spatial-relation VQA, we construct a minimal question pair by swapping the object order, which flips the ground-truth relation. We extract the final-token hidden state at an intermediate layer for each question and compute a \textit{delta vector} as their difference, isolating the relational displacement in embedding space. Aggregated across samples, these vectors summarize the model’s internal spatial representations and enable diagnosing systematic confounds among spatial cues.}
    \label{fig:contrastive_probing}
\end{figure}

\section{Representation Analysis via Contrastive Probing}

Sections~\ref{sec:finding}--\ref{sec:spatialtunnel} established vertical-distance entanglement as a model-intrinsic phenomenon through behavioral evaluation.
We now turn to internal representations to examine how spatial axes are encoded and what distinguishes models that exhibit robust spatial reasoning.

\subsection{Beyond Benchmark Accuracy}
\label{sec:beyond_accuracy}



Behavioral accuracy alone can be a misleading indicator of spatial understanding.
Table~\ref{tab:spatial_benchmarks} reports performance across five spatial reasoning tasks spanning different formats, dimensionalities, and difficulty levels.
Beyond EmbSpatial-Bench and the 3D-spatial split of CV-Bench (CV-3D) used in Table~\ref{tab:real_accuracy}, we additionally include the 2D-spatial split of CV-Bench (CV-2D) and the spatial relationship and relative depth splits of BLINK~\cite{fu2024blink}. Detailed task descriptions are provided in the Appendix~\ref{app:benchmarks_details}.

\begin{table}[t!]
\centering
\caption{\textbf{Performance comparison across spatial understanding benchmarks.} Fine-tuned variants of Molmo, NVILA, and Qwen exhibit inconsistent performance fluctuations depending on the benchmark, while RoboRefer and Qwen3-VL-235B, which show strong representation structure under our probing framework (\Cref{sec:strong_model}), achieve consistently high performance across all evaluated tasks. \textbf{Bold} denotes best. BLINK confidence intervals are reported in Appendix~\ref{app:blink_ci}.}
\label{tab:spatial_benchmarks}
\small
\begin{tabular}{lcccccc}
\toprule
 & \textbf{EmbSpatial} & \textbf{CV-2D} & \multicolumn{2}{c}{\textbf{CV-3D}} & \multicolumn{2}{c}{\textbf{BLINK}} \\
\cmidrule(lr){2-2} \cmidrule(lr){3-3} \cmidrule(lr){4-5} \cmidrule(lr){6-7}
\textbf{Model} & Overall & Relation & Depth & Distance & Rel.\ Depth & Spat.\ Rel. \\
\midrule
Molmo-7B-O-0924 \cite{deitke2025molmo} & 60.7 & 76.3 & 84.5 & 68.5 & 78.2 & 70.6 \\
~~~~+ 80k  & 52.9 & 62.3 & 71.0 & 67.5 & 72.6 & 60.8 \\
~~~~+ 400k & 64.9 & 84.3 & 80.0 & 70.8 & 72.6 & 68.5 \\
~~~~+ 800k & 69.1 & 90.0 & 82.0 & 70.8 & 75.0 & 61.5 \\
~~~~+ 2M   & 74.3 & 93.7 & 87.3 & 81.3 & 71.0 & 69.2 \\
\midrule
NVILA-Lite-2B \cite{liu2025nvila} & 54.0 & 58.6 & 69.2 & 52.3 & 64.5 & 67.1 \\
~~~~+ 80k  & 65.1 & 78.9 & 66.2 & 60.8 & 53.2 & 74.1 \\
~~~~+ 400k & 62.1 & 83.2 & 74.3 & 67.0 & 71.8 & 63.6 \\
~~~~+ 800k & 69.7 & 85.5 & 78.2 & 71.3 & 57.3 & 65.0 \\
~~~~+ 2M   & 69.4 & 91.4 & 93.8 & 87.2 & 70.2 & 62.9 \\
\midrule
RoboRefer-SFT-2B~\cite{zhou2025roborefer} & \textbf{92.0} & \textbf{96.5} & \textbf{95.7} & 90.5 & \textbf{84.7} & 79.7 \\
\midrule
Qwen2.5-VL-3B \cite{Qwen2.5-VL} & 62.3 & 67.4 & 70.3 & 60.2 & 68.6 & 83.9 \\
~~~~+ 80k  & 57.3 & 59.7 & 64.7 & 61.5 & 58.1 & 79.7 \\
~~~~+ 400k & 58.6 & 58.2 & 62.0 & 54.5 & 58.9 & 78.3 \\
~~~~+ 800k & 60.9 & 59.4 & 58.7 & 51.2 & 58.1 & 79.0 \\
~~~~+ 2M   & 65.7 & 68.8 & 58.5 & 52.2 & 53.2 & 78.3 \\
\midrule
Qwen3-VL-235B \cite{bai2025qwen3} & 82.0 & \textbf{96.5} & 93.3 & \textbf{91.0} & \textbf{84.7} & \textbf{90.2} \\
\bottomrule
\end{tabular}
\vspace{-1.5em}
\end{table}

Fine-tuned variants of Molmo, NVILA, and Qwen show inconsistent patterns across benchmarks.
For example, NVILA (2M) achieves $93.8\%$ on CV-3D Depth but only $62.9\%$ on BLINK Spatial Relation, while Qwen (2M) scores $78.3\%$ on BLINK Spatial Relation but drops to $52.2\%$ on CV-3D Distance.
No single accuracy figure reliably indicates how well these models have internalized 3D spatial concepts.
In contrast, RoboRefer-SFT-2B and Qwen3-VL-235B achieve consistently high performance across all benchmarks.
As we show in \Cref{sec:strong_model}, these models also exhibit the most structured internal representations under our probing framework, including high axis coherence, well-separated PCA clusters, and low entanglement.
This suggests that representation quality underlies robust spatial reasoning across benchmarks, motivating us to look beyond accuracy and analyze model-internal representations directly.

\subsection{Contrastive Probing}
\label{sec:probing}
Given an image, we construct a pair of questions that differ only in the ordering of the queried objects, such as swapping \emph{``Is A to the left or right of B?''} with \emph{``Is B to the left or right of A?''}.
Consequently, the ground-truth answer for the swapped query becomes the spatial inverse of the original; for instance, a \textit{left} relationship is inverted to \textit{right}.
For probing, we extract hidden states at a fixed intermediate layer $L^*$ per model family. Let $h_q \in \mathbb{R}^d$ denote the final-token hidden state at layer $L^*$ for question $q$.
For a question pair $(q_1,q_2)$, we define \textbf{delta vector} $\delta = h_{q_2} - h_{q_1}$ as the representation-space displacement induced by the swap.
By repeating this across many images, we obtain a set of delta vectors per spatial category (\eg, \emph{above}, \emph{below}, \emph{far}, \emph{close}, \emph{left}, \emph{right}).
This procedure aims to isolate the latent encoding of spatial directions by neutralizing common visual components. We define two metrics over these delta vectors:

\paragraph{\textbf{Axis coherence.}}
For each spatial axis (horizontal, vertical, distance), we pool the delta vectors from both opposing categories (\eg, \emph{far} and \emph{close} for the distance axis).
To align directions, we negate the deltas from the opposing category so that all vectors point toward the canonical direction:
\begin{equation}
    \tilde{\delta}^{(i)} = \begin{cases}
        \delta^{(i)} & \text{if category is canonical (\eg, \emph{far})} \\
        -\delta^{(i)} & \text{if category is opposite (\eg, \emph{close})}
    \end{cases}
\end{equation}
Axis coherence is the mean pairwise cosine similarity over the sign-corrected set:
\begin{equation}
    \mathrm{Coh}_{\mathrm{axis}} = \frac{2}{N(N-1)} \sum_{i < j} \cos(\tilde{\delta}^{(i)},\; \tilde{\delta}^{(j)}).
\label{eq:coherence}
\end{equation}
High coherence indicates that the model encodes the axis as a stable, consistent direction in representation space.


\paragraph{\textbf{VD-Entanglement Index.}}
To quantify the degree of vertical-distance entanglement at the representation level, we compute the mean delta vector $\mu_c$ for each category $c \in \{\textit{above}, \textit{below}, \textit{far}, \textit{close}\}$ and define the \textbf{VD-Entanglement Index} (VD-EI):
\begin{equation}
\begin{split}
    \mathrm{VD\text{-}EI} = \tfrac{1}{4} \bigl[ 
        &\cos(\mu_{\text{above}}, \mu_{\text{far}}) + \cos(\mu_{\text{below}}, \mu_{\text{close}}) \\
        &- \cos(\mu_{\text{above}}, \mu_{\text{close}}) - \cos(\mu_{\text{below}}, \mu_{\text{far}}) 
    \bigr].
\end{split}
\label{eq:vd_ei}
\end{equation}
The first two terms measure the similarity between perspective-aligned pairs (above$\leftrightarrow$far, below$\leftrightarrow$close); the last two measure perspective-opposing pairs.
A positive value indicates that vertical and distance representations are directionally coupled in the manner predicted by perspective projection; zero indicates independence.
We extract hidden states from EmbSpatial-Bench~\cite{du2024embspatial} images at a fixed intermediate layer per model family, following previous approaches~\cite{gurnee2024language, skean2025layer, chen2025rethinking} (see Appendix~\hyperlink{app_layer}{\textcolor{red}{D.3}} and~\hyperlink{app_layer_robust}{\textcolor{red}{D.4}} for details of layer selection).


\subsection{Distance Coherence and Counter Accuracy}
\label{sec:dist_coherence}

\paragraph{\textbf{Distance coherence is the weakest axis.}}
Across all models and training scales in Table~\ref{tab:coherence}, $\CohD$ is the lowest among the three axes.
Fine-tuning substantially increases vertical coherence (\eg, Molmo: $0.23 \to 0.57$; Qwen: $0.29 \to 0.59$), but \CohD grows by a comparatively smaller margin.

\begin{figure}[t] 
    \centering
    \begin{subfigure}[b]{0.48\linewidth}
        \centering
        \includegraphics[width=\linewidth]{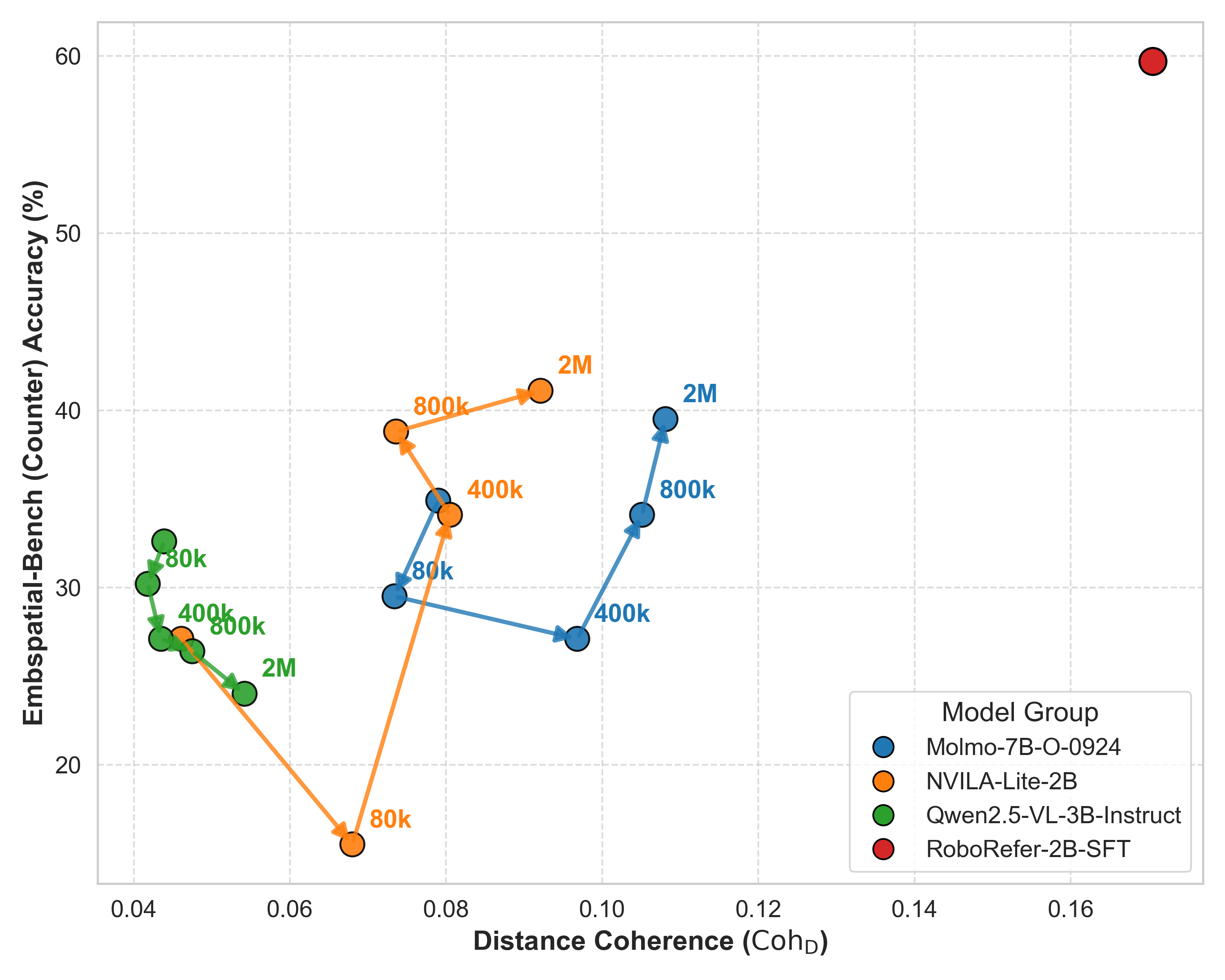}
        \caption{Counter Accuracy vs.\ Distance Coherence}
        \label{fig:dist_coh_vs_counter_acc}
    \end{subfigure}
    \hfill 
    \begin{subfigure}[b]{0.48\linewidth}
        \centering
        \vspace{-1.5em}
        \includegraphics[width=\linewidth]{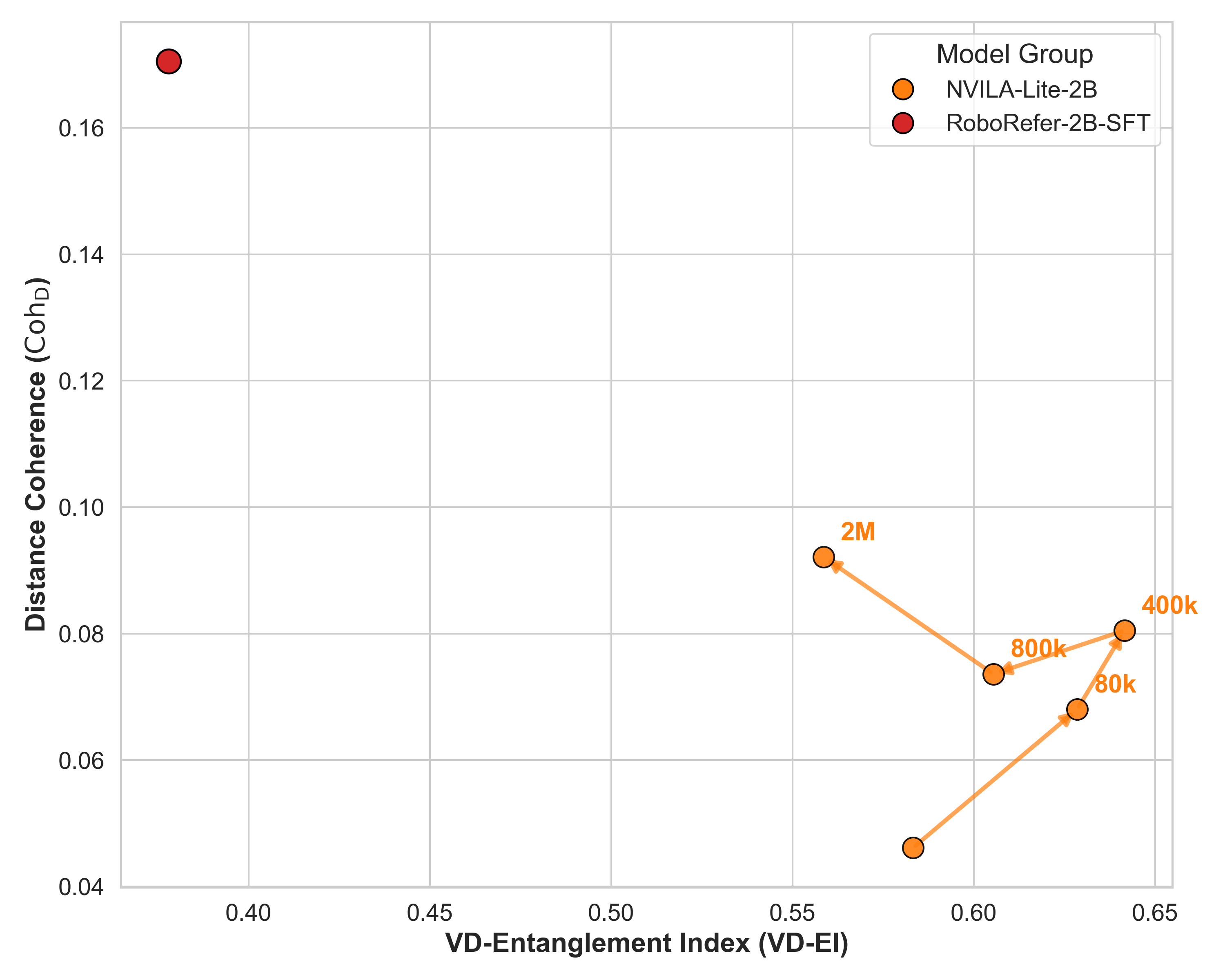}
        \caption{Distance Coherence vs.\ VD-EI}
        \label{fig:dist_coh_vs_ent}
    \end{subfigure}

    \caption{\textbf{Internal probing analysis of spatial representations.} (a) Positive correlation between behavioral accuracy on counter examples and internal distance coherence (\CohD). (b) Comparing distance coherence (\CohD) against geometric entanglement (VD-EI) within the NVILA family; RoboRefer occupies a unique region of high coherence and low entanglement.
    Unlabeled points denote base models, and numeric labels (e.g., 80k) indicate data-mix fine-tuned variants.}
    \label{fig:combined_internal_analysis}
\end{figure}

\paragraph{\textbf{Distance coherence growth accompanies counter accuracy improvement.}}
Figure~\ref{fig:dist_coh_vs_counter_acc} plots $\CohD$ against counter accuracy on EmbSpatial-Bench, the same dataset from which the coherence values are derived.
At early scaling steps (\eg, 80k), counter accuracy sometimes drops before $\CohD$ has meaningfully increased.
However, once spatial fine-tuning data reaches sufficient scale, a consistent pattern emerges across NVILA (from 80k onward) and Molmo (from 400k onward): as $\CohD$ increases, counter accuracy rises in tandem, tracing an upward trajectory in Figure~\ref{fig:dist_coh_vs_counter_acc}.
In contrast, Qwen's $\CohD$ remains nearly flat throughout scaling ($0.043 \to 0.052$), and its counter accuracy declines, widening the consistent-counter gap.
This suggests that as models form a more coherent distance representation through spatial data scaling, they become more robust to the vertical-position shortcut.
Conversely, when distance coherence stagnates, continued scaling does not resolve the entanglement.

\paragraph{\textbf{Cross-domain validity of distance coherence.}}
To examine whether \CohD reflects a reusable representation rather than a benchmark-specific artifact, we measure \CohD on \TUNNEL and compare it against \textit{counter} accuracy on two other benchmarks.
\CohD computed on \TUNNEL correlates with \textit{counter} accuracy on both EmbSpatial-Bench and CV-Bench-3D ($\rho=0.759$ and $0.804$, respectively; both $p<10^{-3}$).
This cross-domain alignment supports the view that \CohD captures predictive signal beyond in-domain computation artifacts (see Appendix~\hyperlink{app_cross}{\textcolor{red}{D.5}} for additional details).

\subsection{What Characterizes Strong Spatial Representations}
\label{sec:strong_model}


\begin{figure}[t]
    \centering
    \includegraphics[width=1.0\linewidth]{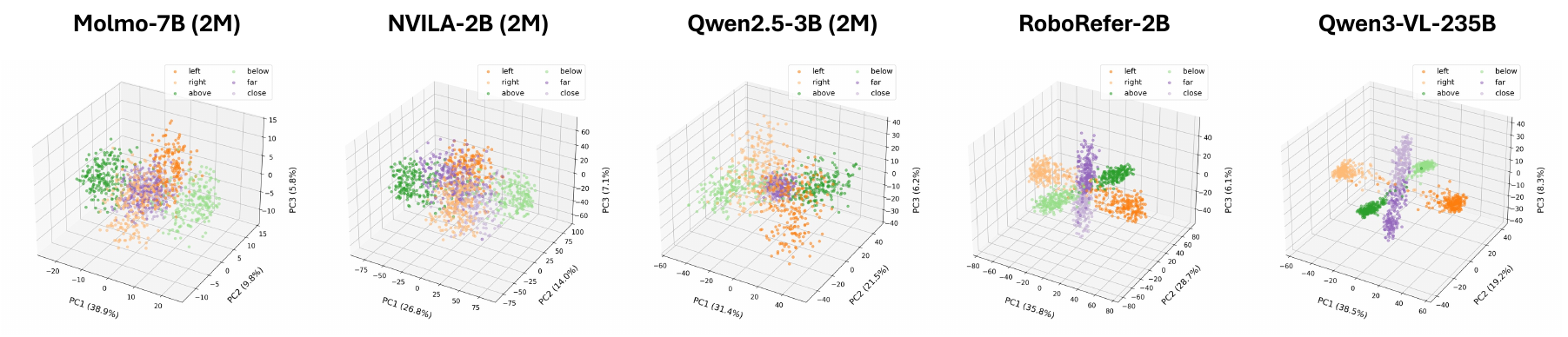}
    \caption{\textbf{PCA of delta vectors across models.} Each point is a delta vector colored by axis (orange: horizontal, green: vertical, purple: distance), with darker/lighter shades distinguishing opposing categories within each axis (\eg, left vs.\ right). Molmo (2M), NVILA (2M), and Qwen (2M) show separation along the horizontal and vertical axes, but distance delta vectors remain poorly distinguished. RoboRefer and Qwen3 exhibit three clearly separated clusters, each aligned with a distinct principal component.}
    \label{fig:pca}
    \vspace{-1em}
\end{figure}


We compare the NVILA-Lite-2B scaling series and RoboRefer-2B-SFT~\cite{zhou2025roborefer}, which share the same base architecture, to identify what representation profile accompanies robust spatial reasoning.
Figure~\ref{fig:pca} visualizes the delta vectors via PCA.
In the base model (\textit{i.e.}, NVILA-Lite-2B), distance delta vectors are collapsed near the origin without forming a distinguishable axis.
Fine-tuning with 2M samples initiates directional spreading, but vertical and distance clusters remain overlapped.
RoboRefer exhibits three cleanly separated clusters, each aligned with a distinct principal component.
Figure~\ref{fig:dist_coh_vs_ent} quantifies this contrast.
The NVILA scaling trajectory yields only marginal gains in $\CohD$ while \VDEI remains high ($0.54$--$0.64$).
RoboRefer occupies a distinct region: the highest $\CohD$ ($0.182$) and the lowest \VDEI ($0.362$) in the family, corresponding to $59.7\%$ counter accuracy on EmbSpatial-Bench versus $41.1\%$ for NVILA (2M).

\begin{wrapfigure}[19]{r}{0.44\linewidth}
\centering
\vspace{-1.1em}
\captionof{table}{%
    \textbf{Axis coherence and VD-Entanglement Index.}
    $\mathrm{Coh}$ measures directional consistency within axes. $\mathrm{Coh}_{\mathrm{D}}$ is consistently the lowest across all models. \textbf{Bold} denotes best.
}
\label{tab:coherence}
\scriptsize
\begin{tabular*}{\linewidth}{@{\extracolsep{\fill}}lcccc@{}}
\toprule
\textbf{Model} & $\mathbf{Coh}_{\mathbf{H}}$ & $\mathbf{Coh}_{\mathbf{V}}$ & $\mathbf{Coh}_{\mathbf{D}}$ & $\mathbf{VD\text{-}EI}$ \\
\midrule
Molmo-7B        & 0.143 & 0.228 & 0.075 & 0.279 \\
~~~~+ 80k       & 0.122 & 0.332 & 0.072 & 0.388 \\
~~~~+ 400k      & 0.236 & 0.597 & 0.096 & 0.459 \\
~~~~+ 800k      & 0.247 & 0.559 & 0.107 & 0.514 \\
~~~~+ 2M        & 0.239 & 0.574 & 0.112 & 0.474 \\
\midrule
NVILA-2B        & 0.323 & 0.289 & 0.052 & 0.539 \\
~~~~+ 80k  & 0.295 & 0.497 & 0.070 & 0.606 \\
~~~~+ 400k & 0.242 & 0.574 & 0.095 & 0.589 \\
~~~~+ 800k & 0.278 & 0.498 & 0.089 & 0.591 \\
~~~~+ 2M   & 0.241 & 0.553 & 0.104 & 0.550 \\
\midrule
RoboRefer-2B    & \textbf{0.649} & \textbf{0.830} & \textbf{0.182} & 0.362 \\
\midrule
Qwen2.5-3B      & 0.367 & 0.293 & 0.043 & 0.457 \\
~~~~+ 80k  & 0.386 & 0.315 & 0.040 & 0.456 \\
~~~~+ 400k & 0.450 & 0.452 & 0.042 & 0.451 \\
~~~~+ 800k & 0.473 & 0.538 & 0.045 & 0.429 \\
~~~~+ 2M   & 0.485 & 0.586 & 0.052 & 0.472 \\
\bottomrule
\end{tabular*}
\vspace{-1em}
\end{wrapfigure}

These results suggest that high $\CohD$, together with low \VDEI as a complementary signal, accompanies robust spatial reasoning across benchmarks.
Within the NVILA scaling series, incremental increases in fine-tuning scale yield only modest gains in $\CohD$.
In contrast, RoboRefer and Qwen3 exhibit well-separated axes and strong overall performance, showing such structure can emerge under substantially richer training regimes.
RoboRefer reflects a different training regime (\eg, additional supervision and much larger training), so we treat it as an illustrative reference rather than attributing its gains to a single factor.
Overall, $\CohD$ can serve as a practical diagnostic for whether training improves spatial representations.

\section{Conclusion}
\label{sec:conclusion}

We introduced a representation-level diagnostic framework that reveals vertical-distance entanglement as a pervasive, model-intrinsic bias across VLM families and scales.
Our analysis shows that models with more structured spatial representations -- characterized by high distance coherence and low VD-Entanglement Index -- not only exhibit stronger counter-heuristic robustness but also achieve higher accuracy across diverse spatial reasoning benchmarks.
To isolate this bias from evaluation-set confounds, we introduced the synthetic benchmark \TUNNEL, which removes perspective-driven correlations present in natural images.
Together, these results demonstrate that representational structure, rather than benchmark accuracy alone, provides a reliable indicator of robust spatial reasoning in vision-language models.

%
%
\bibliographystyle{splncs04}
\bibliography{main}

@String(CVPR  = {IEEE Conf. Comput. Vis. Pattern Recog.})

@String(ICCV  = {Int. Conf. Comput. Vis.})

@String(NeurIPS = {Adv. Neural Inform. Process. Syst.})

@String(CVPR  = {CVPR})

@String(ICCV  = {ICCV})

@String(NeurIPS = {NeurIPS})

@inproceedings{
  zhang2025do,
  title={Do Vision-Language Models Represent Space and How? Evaluating Spatial Frame of Reference under Ambiguities},
  author={Zheyuan Zhang and Fengyuan Hu and Jayjun Lee and Freda Shi and Parisa Kordjamshidi and Joyce Chai and Ziqiao Ma},
  booktitle={The Thirteenth International Conference on Learning Representations},
  year={2025},
  url={https://openreview.net/forum?id=84pDoCD4lH}
}

@inproceedings{song2025robospatial,
  title={Robospatial: Teaching spatial understanding to 2d and 3d vision-language models for robotics},
  author={Song, Chan Hee and Blukis, Valts and Tremblay, Jonathan and Tyree, Stephen and Su, Yu and Birchfield, Stan},
  booktitle={Proceedings of the Computer Vision and Pattern Recognition Conference},
  pages={15768--15780},
  year={2025}
}

@inproceedings{fu2024blink,
  title={Blink: Multimodal large language models can see but not perceive},
  author={Fu, Xingyu and Hu, Yushi and Li, Bangzheng and Feng, Yu and Wang, Haoyu and Lin, Xudong and Roth, Dan and Smith, Noah A and Ma, Wei-Chiu and Krishna, Ranjay},
  booktitle={European Conference on Computer Vision},
  pages={148--166},
  year={2024},
  organization={Springer}
}

@inproceedings{du2024embspatial,
  title={Embspatial-bench: Benchmarking spatial understanding for embodied tasks with large vision-language models},
  author={Du, Mengfei and Wu, Binhao and Li, Zejun and Huang, Xuan-Jing and Wei, Zhongyu},
  booktitle={Proceedings of the 62nd Annual Meeting of the Association for Computational Linguistics (Volume 2: Short Papers)},
  pages={346--355},
  year={2024}
}

@inproceedings{
ray2025sat,
title={{SAT}: Dynamic Spatial Aptitude Training for Multimodal Language Models},
author={Arijit Ray and Jiafei Duan and Ellis L Brown II and Reuben Tan and Dina Bashkirova and Rose Hendrix and Kiana Ehsani and Aniruddha Kembhavi and Bryan A. Plummer and Ranjay Krishna and Kuo-Hao Zeng and Kate Saenko},
booktitle={Second Conference on Language Modeling},
year={2025},
url={https://openreview.net/forum?id=DW8U8ZWa1U}
}

@inproceedings{
zhou2025roborefer,
title={RoboRefer: Towards Spatial Referring with Reasoning in Vision-Language Models for Robotics},
author={Enshen Zhou and Jingkun An and Cheng Chi and Yi Han and Shanyu Rong and Chi Zhang and Pengwei Wang and Zhongyuan Wang and Tiejun Huang and Lu Sheng and Shanghang Zhang},
booktitle={The Thirty-ninth Annual Conference on Neural Information Processing Systems},
year={2025},
url={https://openreview.net/forum?id=OGxalNUHbJ}
}

@article{team2025gemini,
  title={Gemini robotics: Bringing ai into the physical world},
  author={Team, Gemini Robotics and Abeyruwan, Saminda and Ainslie, Joshua and Alayrac, Jean-Baptiste and Arenas, Montserrat Gonzalez and Armstrong, Travis and Balakrishna, Ashwin and Baruch, Robert and Bauza, Maria and Blokzijl, Michiel and others},
  journal={arXiv preprint arXiv:2503.20020},
  year={2025}
}

@misc{zhang2025mllmsstrugglespatialunderstanding,
      title={Why Do MLLMs Struggle with Spatial Understanding? A Systematic Analysis from Data to Architecture}, 
      author={Wanyue Zhang and Yibin Huang and Yangbin Xu and JingJing Huang and Helu Zhi and Shuo Ren and Wang Xu and Jiajun Zhang},
      year={2025},
      eprint={2509.02359},
      archivePrefix={arXiv},
      primaryClass={cs.CV},
      url={https://arxiv.org/abs/2509.02359}, 
}

@inproceedings{
chen2025why,
title={Why Is Spatial Reasoning Hard for {VLM}s? An Attention Mechanism Perspective on Focus Areas},
author={Shiqi Chen and Tongyao Zhu and Ruochen Zhou and Jinghan Zhang and Siyang Gao and Juan Carlos Niebles and Mor Geva and Junxian He and Jiajun Wu and Manling Li},
booktitle={Forty-second International Conference on Machine Learning},
year={2025},
url={https://openreview.net/forum?id=k7vcuqLK4X}
}

@inproceedings{
tong2024cambrian,
title={Cambrian-1: A Fully Open, Vision-Centric Exploration of Multimodal {LLM}s},
author={Shengbang Tong and Ellis L Brown II and Penghao Wu and Sanghyun Woo and ADITHYA JAIRAM IYER and Sai Charitha Akula and Shusheng Yang and Jihan Yang and Manoj Middepogu and Ziteng Wang and Xichen Pan and Rob Fergus and Yann LeCun and Saining Xie},
booktitle={The Thirty-eighth Annual Conference on Neural Information Processing Systems},
year={2024},
url={https://openreview.net/forum?id=Vi8AepAXGy}
}

@inproceedings{kamath2023whatsup,
  title     = {What's ``Up'' with Vision-Language Models? Investigating Their Struggle with Spatial Reasoning},
  author    = {Kamath, Amita and Hessel, Jack and Chang, Kai-Wei},
  booktitle = {Proceedings of the Conference on Empirical Methods in Natural Language Processing (EMNLP)},
  year      = {2023}
}

@inproceedings{chen2024spatialvlm,
  title={Spatialvlm: Endowing vision-language models with spatial reasoning capabilities},
  author={Chen, Boyuan and Xu, Zhuo and Kirmani, Sean and Ichter, Brain and Sadigh, Dorsa and Guibas, Leonidas and Xia, Fei},
  booktitle={Proceedings of the IEEE/CVF Conference on Computer Vision and Pattern Recognition},
  pages={14455--14465},
  year={2024}
}

@manual{blender,
  title        = {Blender - a 3D modelling and rendering package},
  author       = {{Blender Online Community}},
  organization = {Blender Foundation},
  year         = {2025},
  url          = {https://www.blender.org/},
  urldate      = {2026-03-02}
}

@inproceedings{danier2025depthcues,
  title={DepthCues: Evaluating monocular depth perception in large vision models},
  author={Danier, Duolikun and Ayg{\"u}n, Mehmet and Li, Changjian and Bilen, Hakan and Mac Aodha, Oisin},
  booktitle={Proceedings of the Computer Vision and Pattern Recognition Conference},
  pages={20049--20059},
  year={2025}
}

@inproceedings{wang2025spatial457,
  title={Spatial457: A diagnostic benchmark for 6d spatial reasoning of large mutimodal models},
  author={Wang, Xingrui and Ma, Wufei and Zhang, Tiezheng and de Melo, Celso M and Chen, Jieneng and Yuille, Alan},
  booktitle={Proceedings of the Computer Vision and Pattern Recognition Conference},
  pages={24669--24679},
  year={2025}
}

@inproceedings{sheta2025behavioral,
  title={From Behavioral Performance to Internal Competence: Interpreting Vision-Language Models with VLM-LENS},
  author={Sheta, Hala and Huang, Eric Haoran and Wu, Shuyu and Alenabi, Ilia and Hong, Jiajun and Lin, Ryker and Ning, Ruoxi and Wei, Daniel and Yang, Jialin and Zhou, Jiawei and others},
  booktitle={Proceedings of the 2025 Conference on Empirical Methods in Natural Language Processing: System Demonstrations},
  pages={886--895},
  year={2025}
}

@article{li2025spatialforcing,
  title={Spatial forcing: Implicit spatial representation alignment for vision-language-action model},
  author={Li, Fuhao and Song, Wenxuan and Zhao, Han and Wang, Jingbo and Ding, Pengxiang and Wang, Donglin and Zeng, Long and Li, Haoang},
  journal={arXiv preprint arXiv:2510.12276},
  year={2025}
}

@article{kang2026linearprobing,
  title={Linear Mechanisms for Spatiotemporal Reasoning in Vision Language Models},
  author={Kang, Raphi and Chen, Hongqiao and Gkioxari, Georgia and Perona, Pietro},
  journal={arXiv preprint arXiv:2601.12626},
  year={2026}
}

@inproceedings{hu2023prompting,
  title={Prompting is not a substitute for probability measurements in large language models},
  author={Hu, Jennifer and Levy, Roger},
  booktitle={Proceedings of the 2023 Conference on Empirical Methods in Natural Language Processing},
  pages={5040--5060},
  year={2023}
}

@inproceedings{wang2025logical,
  title={Logical forms complement probability in understanding language model (and human) performance},
  author={Wang, Yixuan and Shi, Freda},
  booktitle={Proceedings of the 63rd Annual Meeting of the Association for Computational Linguistics (Volume 1: Long Papers)},
  pages={16862--16877},
  year={2025}
}

@inproceedings{deitke2025molmo,
  title={Molmo and pixmo: Open weights and open data for state-of-the-art vision-language models},
  author={Deitke, Matt and Clark, Christopher and Lee, Sangho and Tripathi, Rohun and Yang, Yue and Park, Jae Sung and Salehi, Mohammadreza and Muennighoff, Niklas and Lo, Kyle and Soldaini, Luca and others},
  booktitle={Proceedings of the Computer Vision and Pattern Recognition Conference},
  pages={91--104},
  year={2025}
}

@inproceedings{liu2025nvila,
  title={Nvila: Efficient frontier visual language models},
  author={Liu, Zhijian and Zhu, Ligeng and Shi, Baifeng and Zhang, Zhuoyang and Lou, Yuming and Yang, Shang and Xi, Haocheng and Cao, Shiyi and Gu, Yuxian and Li, Dacheng and others},
  booktitle={Proceedings of the IEEE/CVF Conference on Computer Vision and Pattern Recognition},
  pages={4122--4134},
  year={2025}
}

@article{Qwen2.5-VL,
  title={Qwen2.5-VL Technical Report},
  author={Bai, Shuai and Chen, Keqin and Liu, Xuejing and Wang, Jialin and Ge, Wenbin and Song, Sibo and Dang, Kai and Wang, Peng and Wang, Shijie and Tang, Jun and Zhong, Humen and Zhu, Yuanzhi and Yang, Mingkun and Li, Zhaohai and Wan, Jianqiang and Wang, Pengfei and Ding, Wei and Fu, Zheren and Xu, Yiheng and Ye, Jiabo and Zhang, Xi and Xie, Tianbao and Cheng, Zesen and Zhang, Hang and Yang, Zhibo and Xu, Haiyang and Lin, Junyang},
  journal={arXiv preprint arXiv:2502.13923},
  year={2025}
}

@article{bai2025qwen3,
  title={Qwen3-vl technical report},
  author={Bai, Shuai and Cai, Yuxuan and Chen, Ruizhe and Chen, Keqin and Chen, Xionghui and Cheng, Zesen and Deng, Lianghao and Ding, Wei and Gao, Chang and Ge, Chunjiang and others},
  journal={arXiv preprint arXiv:2511.21631},
  year={2025}
}

@inproceedings{gurnee2024language,
 author = {Gurnee, Wes and Tegmark, Max },
 booktitle = {International Conference on Learning Representations},
 editor = {B. Kim and Y. Yue and S. Chaudhuri and K. Fragkiadaki and M. Khan and Y. Sun},
 pages = {2483--2503},
 title = {Language Models Represent Space and Time},
 url = {https://proceedings.iclr.cc/paper_files/paper/2024/file/0a6059857ae5c82ea9726ee9282a7145-Paper-Conference.pdf},
 volume = {2024},
 year = {2024}
}

@inproceedings{skean2025layer,
title={Layer by Layer: Uncovering Hidden Representations in Language Models},
author={Oscar Skean and Md Rifat Arefin and Dan Zhao and Niket Nikul Patel and Jalal Naghiyev and Yann LeCun and Ravid Shwartz-Ziv},
booktitle={Forty-second International Conference on Machine Learning},
year={2025},
url={https://openreview.net/forum?id=WGXb7UdvTX}
}

@article{chen2025rethinking,
  title={Rethinking visual layer selection in multimodal llms},
  author={Chen, Haoran and Lin, Junyan and Chen, Xinhao and Fan, Yue and Jin, Xin and Su, Hui and Dong, Jianfeng and Fu, Jinlan and Shen, Xiaoyu},
  journal={arXiv e-prints},
  pages={arXiv--2504},
  year={2025}
}

@article{zhang2025flatland,
  title={From flatland to space: Teaching vision-language models to perceive and reason in 3d},
  author={Zhang, Jiahui and Chen, Yurui and Zhou, Yanpeng and Xu, Yueming and Huang, Ze and Mei, Jilin and Chen, Junhui and Yuan, Yu-Jie and Cai, Xinyue and Huang, Guowei and others},
  journal={arXiv preprint arXiv:2503.22976},
  year={2025}
}

@article{deshpande2025graspmolmo,
  title={Graspmolmo: Generalizable task-oriented grasping via large-scale synthetic data generation},
  author={Deshpande, Abhay and Deng, Yuquan and Ray, Arijit and Salvador, Jordi and Han, Winson and Duan, Jiafei and Zeng, Kuo-Hao and Zhu, Yuke and Krishna, Ranjay and Hendrix, Rose},
  journal={arXiv preprint arXiv:2505.13441},
  year={2025}
}

@inproceedings{chen2026spacetools,
  title     = {{SpaceTools}: Tool-Augmented Spatial Reasoning via Double Interactive {RL}},
  author    = {Chen, Siyi and Uy, Mikaela Angelina and Song, Chan Hee and Ladhak, Faisal and Murali, Adithyavairavan and Qu, Qing and Birchfield, Stan and Blukis, Valts and Tremblay, Jonathan},
  booktitle = {Proceedings of the IEEE/CVF Conference on Computer Vision and Pattern Recognition (CVPR)},
  year      = {2026},
  note      = {to appear},
  url       = {https://arxiv.org/abs/2512.04069}
}

@InProceedings{kim24openvla,
  title = 	 {OpenVLA: An Open-Source Vision-Language-Action Model},
  author =       {Kim, Moo Jin and Pertsch, Karl and Karamcheti, Siddharth and Xiao, Ted and Balakrishna, Ashwin and Nair, Suraj and Rafailov, Rafael and Foster, Ethan P and Sanketi, Pannag R and Vuong, Quan and Kollar, Thomas and Burchfiel, Benjamin and Tedrake, Russ and Sadigh, Dorsa and Levine, Sergey and Liang, Percy and Finn, Chelsea},
  booktitle = {Proceedings of the 8th Conference on Robot Learning (CoRL)},
  pages = 	 {2679--2713},
  year = 	 {2025},
  editor = 	 {Agrawal, Pulkit and Kroemer, Oliver and Burgard, Wolfram},
  volume = 	 {270},
  series = 	 {Proceedings of Machine Learning Research},
  month = 	 {06--09 Nov},
  publisher =    {PMLR},
  url = 	 {https://proceedings.mlr.press/v270/kim25c.html}
}

@misc{nvidia2025gr00tn1openfoundation,
      title={GR00T N1: An Open Foundation Model for Generalist Humanoid Robots}, 
      author={NVIDIA and : and Johan Bjorck and Fernando Castañeda and Nikita Cherniadev and Xingye Da and Runyu Ding and Linxi "Jim" Fan and Yu Fang and Dieter Fox and Fengyuan Hu and Spencer Huang and Joel Jang and Zhenyu Jiang and Jan Kautz and Kaushil Kundalia and Lawrence Lao and Zhiqi Li and Zongyu Lin and Kevin Lin and Guilin Liu and Edith Llontop and Loic Magne and Ajay Mandlekar and Avnish Narayan and Soroush Nasiriany and Scott Reed and You Liang Tan and Guanzhi Wang and Zu Wang and Jing Wang and Qi Wang and Jiannan Xiang and Yuqi Xie and Yinzhen Xu and Zhenjia Xu and Seonghyeon Ye and Zhiding Yu and Ao Zhang and Hao Zhang and Yizhou Zhao and Ruijie Zheng and Yuke Zhu},
      year={2025},
      eprint={2503.14734},
      archivePrefix={arXiv},
      primaryClass={cs.RO},
      url={https://arxiv.org/abs/2503.14734}, 
}

@misc{intelligence2025pi05visionlanguageactionmodelopenworld,
      title={$\pi_{0.5}$: a Vision-Language-Action Model with Open-World Generalization}, 
      author={Physical Intelligence and Kevin Black and Noah Brown and James Darpinian and Karan Dhabalia and Danny Driess and Adnan Esmail and Michael Equi and Chelsea Finn and Niccolo Fusai and Manuel Y. Galliker and Dibya Ghosh and Lachy Groom and Karol Hausman and Brian Ichter and Szymon Jakubczak and Tim Jones and Liyiming Ke and Devin LeBlanc and Sergey Levine and Adrian Li-Bell and Mohith Mothukuri and Suraj Nair and Karl Pertsch and Allen Z. Ren and Lucy Xiaoyang Shi and Laura Smith and Jost Tobias Springenberg and Kyle Stachowicz and James Tanner and Quan Vuong and Homer Walke and Anna Walling and Haohuan Wang and Lili Yu and Ury Zhilinsky},
      year={2025},
      eprint={2504.16054},
      archivePrefix={arXiv},
      primaryClass={cs.LG},
      url={https://arxiv.org/abs/2504.16054}, 
}

@article{tan2026robobrain25depthsight,
    title={RoboBrain 2.5: Depth in Sight, Time in Mind},
    author={Tan, Huajie and Zhou, Enshen and Li, Zhiyu and Xu, Yijie and Ji, Yuheng and Chen, Xiansheng and Chi, Cheng and Wang, Pengwei and Jia, Huizhu and Ao, Yulong and Cao, Mingyu and Chen, Sixiang and Li, Zhe and Liu, Mengzhen and Wang, Zixiao and Rong, Shanyu and Lyu, Yaoxu and Zhao, Zhongxia and Co, Peterson and Li, Yibo and Han, Yi and Xie, Shaoxuan and Yao, Guocai and Wang, Songjing and Zhang, Leiduo and Yang, Xi and Jiao, Yance and Shi, Donghai and Xie, Kunchang and Nie, Shaokai and Men, Chunlei and Lin, Yonghua and Wang, Zhongyuan and Huang, Tiejun and Zhang, Shanghang},
    journal={arXiv preprint arXiv:2601.14352},
    year={2026}
}

@inproceedings{llm-planner,
  title={Llm-planner: Few-shot grounded planning for embodied agents with large language models},
  author={Song, Chan Hee and Wu, Jiaman and Washington, Clayton and Sadler, Brian M and Chao, Wei-Lun and Su, Yu},
  booktitle={ICCV},
  year={2023}
}

@article{ahn2022can,
  title={Do as i can, not as i say: Grounding language in robotic affordances},
  author={Ahn, Michael and Brohan, Anthony and Brown, Noah and Chebotar, Yevgen and Cortes, Omar and David, Byron and Finn, Chelsea and Fu, Chuyuan and Gopalakrishnan, Keerthana and Hausman, Karol and others},
  journal={arXiv preprint arXiv:2204.01691},
  year={2022}
}

@inproceedings{singh2023progprompt,
  author={Singh, Ishika and Blukis, Valts and Mousavian, Arsalan and Goyal, Ankit and Xu, Danfei and Tremblay, Jonathan and Fox, Dieter and Thomason, Jesse and Garg, Animesh},
  booktitle={2023 IEEE International Conference on Robotics and Automation (ICRA)}, 
  title={ProgPrompt: Generating Situated Robot Task Plans using Large Language Models}, 
  year={2023},
  volume={},
  number={},
  pages={11523-11530},
  doi={10.1109/ICRA48891.2023.10161317}
}

@inproceedings{cheng2024spatialrgpt,
          title={SpatialRGPT: Grounded Spatial Reasoning in Vision-Language Models},
          author={Cheng, An-Chieh and Yin, Hongxu and Fu, Yang and Guo, Qiushan and Yang, Ruihan and Kautz, Jan and Wang, Xiaolong and Liu, Sifei},
          booktitle={Advances in Neural Information Processing Systems (NeurIPS)},
          year={2024}
  }

@misc{singh2025openaigpt5card,
      title={OpenAI GPT-5 System Card}, 
      author={{OpenAI}},
      year={2025},
      eprint={2601.03267},
      archivePrefix={arXiv},
      primaryClass={cs.CL},
      url={https://arxiv.org/abs/2601.03267}, 
}

@techreport{anthropic2025claude,
  title        = {Claude Opus 4 \& Claude Sonnet 4 System Card},
  author       = {{Anthropic}},
  year         = {2025},
  month        = {May},
  institution  = {Anthropic},
  url          = {https://www-cdn.anthropic.com/4263b940cabb546aa0e3283f35b686f4f3b2ff47.pdf}
}

@misc{comanici2025gemini25pushingfrontier,
      title={Gemini 2.5: Pushing the Frontier with Advanced Reasoning, Multimodality, Long Context, and Next Generation Agentic Capabilities}, 
      author={{Google}},
      year={2025},
      eprint={2507.06261},
      archivePrefix={arXiv},
      primaryClass={cs.CL},
      url={https://arxiv.org/abs/2507.06261}, 
}

@article{LLaVA-1.5,
  title={Improved baselines with visual instruction tuning},
  author={Liu, Haotian and Li, Chunyuan and Li, Yuheng and Lee, Yong Jae},
  journal={arXiv preprint arXiv:2310.03744},
  year={2023}
}

@article{kuznetsova2020open,
  title={The open images dataset v4: Unified image classification, object detection, and visual relationship detection at scale},
  author={Kuznetsova, Alina and Rom, Hassan and Alldrin, Neil and Uijlings, Jasper and Krasin, Ivan and Pont-Tuset, Jordi and Kamali, Shahab and Popov, Stefan and Malloci, Matteo and Kolesnikov, Alexander and others},
  journal={International journal of computer vision},
  volume={128},
  number={7},
  pages={1956--1981},
  year={2020},
  publisher={Springer}
}

@inproceedings{lazarow2025cubify,
  title={Cubify anything: Scaling indoor 3d object detection},
  author={Lazarow, Justin and Griffiths, David and Kohavi, Gefen and Crespo, Francisco and Dehghan, Afshin},
  booktitle={Proceedings of the Computer Vision and Pattern Recognition Conference},
  pages={22225--22233},
  year={2025}
}

@inproceedings{dai2017scannet,
  title={Scannet: Richly-annotated 3d reconstructions of indoor scenes},
  author={Dai, Angela and Chang, Angel X and Savva, Manolis and Halber, Maciej and Funkhouser, Thomas and Nie{\ss}ner, Matthias},
  booktitle={Proceedings of the IEEE conference on computer vision and pattern recognition},
  pages={5828--5839},
  year={2017}
}

@inproceedings{yeshwanth2023scannet++,
  title={Scannet++: A high-fidelity dataset of 3d indoor scenes},
  author={Yeshwanth, Chandan and Liu, Yueh-Cheng and Nie{\ss}ner, Matthias and Dai, Angela},
  booktitle={Proceedings of the IEEE/CVF International Conference on Computer Vision},
  pages={12--22},
  year={2023}
}

@inproceedings{zheng2020structured3d,
  title={Structured3d: A large photo-realistic dataset for structured 3d modeling},
  author={Zheng, Jia and Zhang, Junfei and Li, Jing and Tang, Rui and Gao, Shenghua and Zhou, Zihan},
  booktitle={European Conference on Computer Vision},
  pages={519--535},
  year={2020},
  organization={Springer}
}

@inproceedings{raistrick2024infinigen,
  title={Infinigen indoors: Photorealistic indoor scenes using procedural generation},
  author={Raistrick, Alexander and Mei, Lingjie and Kayan, Karhan and Yan, David and Zuo, Yiming and Han, Beining and Wen, Hongyu and Parakh, Meenal and Alexandropoulos, Stamatis and Lipson, Lahav and others},
  booktitle={Proceedings of the IEEE/CVF Conference on Computer Vision and Pattern Recognition},
  pages={21783--21794},
  year={2024}
}

@inproceedings{deitke2023objaverse,
  title={Objaverse: A universe of annotated 3d objects},
  author={Deitke, Matt and Schwenk, Dustin and Salvador, Jordi and Weihs, Luca and Michel, Oscar and VanderBilt, Eli and Schmidt, Ludwig and Ehsani, Kiana and Kembhavi, Aniruddha and Farhadi, Ali},
  booktitle={Proceedings of the IEEE/CVF conference on computer vision and pattern recognition},
  pages={13142--13153},
  year={2023}
}

@article{chang2015shapenet,
  title={Shapenet: An information-rich 3d model repository},
  author={Chang, Angel X and Funkhouser, Thomas and Guibas, Leonidas and Hanrahan, Pat and Huang, Qixing and Li, Zimo and Savarese, Silvio and Savva, Manolis and Song, Shuran and Su, Hao and others},
  journal={arXiv preprint arXiv:1512.03012},
  year={2015}
}

@inproceedings{savva2015semantically,
  title={Semantically-enriched 3D models for common-sense knowledge},
  author={Savva, Manolis and Chang, Angel X and Hanrahan, Pat},
  booktitle={Proceedings of the IEEE Conference on Computer Vision and Pattern Recognition Workshops},
  pages={24--31},
  year={2015}
}

@inproceedings{eppner2021acronym,
  title={Acronym: A large-scale grasp dataset based on simulation},
  author={Eppner, Clemens and Mousavian, Arsalan and Fox, Dieter},
  booktitle={2021 IEEE International Conference on Robotics and Automation (ICRA)},
  pages={6222--6227},
  year={2021},
  organization={IEEE}
}

@article{eppner2025scene_synthesizer,
  title={scene\_synthesizer: A Python Library for Procedural Scene Generation in Robot Manipulation},
  author={Eppner, Clemens and Murali, Adithyavairavan and Garrett, Caelan and O'Flaherty, Rowland and Hermans, Tucker and Yang, Wei and Fox, Dieter},
  journal={Journal of Open Source Software},
  volume={10},
  number={105},
  pages={7561},
  year={2025}
}

@inproceedings{lin2014microsoft,
  title={Microsoft coco: Common objects in context},
  author={Lin, Tsung-Yi and Maire, Michael and Belongie, Serge and Hays, James and Perona, Pietro and Ramanan, Deva and Doll{\'a}r, Piotr and Zitnick, C Lawrence},
  booktitle={European conference on computer vision},
  pages={740--755},
  year={2014},
  organization={Springer}
}

@article{krishna2017visual,
  title={Visual genome: Connecting language and vision using crowdsourced dense image annotations},
  author={Krishna, Ranjay and Zhu, Yuke and Groth, Oliver and Johnson, Justin and Hata, Kenji and Kravitz, Joshua and Chen, Stephanie and Kalantidis, Yannis and Li, Li-Jia and Shamma, David A and others},
  journal={International journal of computer vision},
  volume={123},
  number={1},
  pages={32--73},
  year={2017},
  publisher={Springer}
}

@article{chang2017matterport3d,
  title={Matterport3d: Learning from rgb-d data in indoor environments},
  author={Chang, Angel and Dai, Angela and Funkhouser, Thomas and Halber, Maciej and Niessner, Matthias and Savva, Manolis and Song, Shuran and Zeng, Andy and Zhang, Yinda},
  journal={arXiv preprint arXiv:1709.06158},
  year={2017}
}

@article{kolve2017ai2,
  title={Ai2-thor: An interactive 3d environment for visual ai},
  author={Kolve, Eric and Mottaghi, Roozbeh and Han, Winson and VanderBilt, Eli and Weihs, Luca and Herrasti, Alvaro and Deitke, Matt and Ehsani, Kiana and Gordon, Daniel and Zhu, Yuke and others},
  journal={arXiv preprint arXiv:1712.05474},
  year={2017}
}

@article{zhou2019semantic,
  title={Semantic understanding of scenes through the ade20k dataset},
  author={Zhou, Bolei and Zhao, Hang and Puig, Xavier and Xiao, Tete and Fidler, Sanja and Barriuso, Adela and Torralba, Antonio},
  journal={International journal of computer vision},
  volume={127},
  number={3},
  pages={302--321},
  year={2019},
  publisher={Springer}
}

@inproceedings{brazil2023omni3d,
  title={Omni3d: A large benchmark and model for 3d object detection in the wild},
  author={Brazil, Garrick and Kumar, Abhinav and Straub, Julian and Ravi, Nikhila and Johnson, Justin and Gkioxari, Georgia},
  booktitle={Proceedings of the IEEE/CVF conference on computer vision and pattern recognition},
  pages={13154--13164},
  year={2023}
}

@article{deitke2022proc,
title={ProcTHOR: Large-Scale Embodied AI Using Procedural Generation},
  author={Deitke, Matt and VanderBilt, Eli and Herrasti, Alvaro and Weihs, Luca and Ehsani, Kiana and Salvador, Jordi and Han, Winson and Kolve, Eric and Kembhavi, Aniruddha and Mottaghi, Roozbeh},
  journal={Advances in Neural Information Processing Systems},
  volume={35},
  pages={5982--5994},
  year={2022}
}

@book{szeliski2022computer,
  title={Computer vision: algorithms and applications},
  author={Szeliski, Richard},
  year={2022},
  publisher={Springer Nature}
}

@article{hata2015cs231a,
  title={Cs231a course notes 1: Camera models},
  author={Hata, Kenji and Savarese, Silvio},
  journal={url: https://web. stanford. edu/class/cs231a/course\_ notes/01-camera-models. pdf},
  year={2015}
}

@book{hoiem2022representations,
  title={Representations and techniques for 3D object recognition and scene interpretation},
  author={Hoiem, Derek and Savarese, Silvio},
  year={2022},
  publisher={Springer Nature}
}

@article{touvron2023llama,
  title={Llama 2: Open foundation and fine-tuned chat models},
  author={Touvron, Hugo and Martin, Louis and Stone, Kevin and Albert, Peter and Almahairi, Amjad and Babaei, Yasmine and Bashlykov, Nikolay and Batra, Soumya and Bhargava, Prajjwal and Bhosale, Shruti and others},
  journal={arXiv preprint arXiv:2307.09288},
  year={2023}
}

@inproceedings{radford2021learning,
  title={Learning transferable visual models from natural language supervision},
  author={Radford, Alec and Kim, Jong Wook and Hallacy, Chris and Ramesh, Aditya and Goh, Gabriel and Agarwal, Sandhini and Sastry, Girish and Askell, Amanda and Mishkin, Pamela and Clark, Jack and others},
  booktitle={International conference on machine learning},
  pages={8748--8763},
  year={2021},
  organization={PmLR}
}

\clearpage
\appendix

\section*{Appendices}

The following appendices provide supplementary material that complements the main text.

\begin{itemize}
  \item \hyperref[anchor:appA]{\textcolor{blue}{Appendix A}} Ground-Plane Geometry and Vertical Image Position.
  \item \hyperref[anchor:appB]{\textcolor{blue}{Appendix B}} Additional Details on Experiment Setup.
  \item \hyperref[anchor:appC]{\textcolor{blue}{Appendix C}} Additional Details on \TUNNEL.
  \item \hyperref[anchor:appD]{\textcolor{blue}{Appendix D}} Additional Details on Contrastive Probing.
\end{itemize}

\phantomsection
\label{anchor:appA}
\section{Ground-Plane Geometry and Vertical Image Position}
\label{app:ground-plane-geometry}

We derive how perspective projection on a flat ground plane induces a monotonic relationship between an object's depth and its vertical image coordinate under a standard pinhole camera model~\cite{szeliski2022computer, hata2015cs231a}.

\paragraph{\textbf{Setup.}}

Consider a pinhole camera with focal length $f$, whose optical axis is aligned with the world $Z$-axis, and whose center is at height $H_c > 0$ above a horizontal ground plane~\cite{hata2015cs231a}.\footnote{We assume zero camera tilt, square pixels, zero skew, and no lens distortion for simplicity.}

World coordinates are $(X, Y, Z)$, with the ground plane defined as $Y = 0$, and the camera is located at $(0, H_c, 0)$, looking along the positive $Z$-axis. 
In the camera coordinate system, a point on the ground plane has coordinates
\[
(X_c, Y_c, Z_c) = (X, -H_c, Z),
\]
where $Z > 0$ denotes the depth of the point.

\paragraph{\textbf{Perspective projection.}}
Under the pinhole camera model, the projection of $(X_c, Y_c, Z_c)$ onto the image plane at distance $f$ along the $Z_c$-axis is given by~\cite{hata2015cs231a}
\[
u = f \frac{X_c}{Z_c}, \quad v = f \frac{Y_c}{Z_c}.
\]
Substituting the ground-plane coordinates $(X_c, Y_c, Z_c) = (X, -H_c, Z)$ yields
\[
u = f \frac{X}{Z}, \quad v = - f \frac{H_c}{Z}.
\]
Thus, for points on the ground plane with fixed camera height $H_c$, the vertical image coordinate satisfies
\[
v(Z) = - f \frac{H_c}{Z}.
\]


\paragraph{\textbf{Depth--height relationship.}}
From the expression above, the magnitude of the vertical coordinate obeys
\[
\lvert v(Z) \rvert \propto \frac{1}{Z},
\]
so that increasing depth $Z$ decreases the magnitude $\lvert v \rvert$.
Adopting the standard image convention that the $v$-axis increases downward,
we define the image-frame coordinate as
\[
v_{\mathrm{img}}(Z) = -v(Z) = \frac{f H_c}{Z} > 0,
\]
which confirms that ground-plane points appear \emph{below} the principal point.
Under the zero-tilt assumption, the horizon coincides with the principal point at $v_{\mathrm{img}} = 0$~\cite{hoiem2022representations},
and as $Z \to \infty$, we have $v_{\mathrm{img}}(Z) \to 0^{+}$, meaning that
points farther along the ground plane project closer to the horizon line and
therefore appear \emph{higher} in the image.
Consequently, for objects resting on a common ground plane, greater depth
corresponds to a higher vertical position in the image, which is precisely
the classical \emph{elevation} monocular depth cue exploited in both human
perception and recent depth-cue benchmarks~\cite{danier2025depthcues}.

\phantomsection
\label{anchor:appB}
\section{Additional Details on Experiment Setup}
\label{app:experiment_details}
In this section, we detail the models, training data sources, data mix composition, and benchmarks used in the experiments described in Section~\ref{sec:finding}.

\subsection{Models}
We conduct experiments on the following vision-language models, each capable of spatial reasoning.
\begin{itemize}
    \item \textbf{Molmo-7B-O-0924}~\cite{deitke2025molmo}: 
    Molmo-7B-O-0924 is a 7B-parameter open vision-language model from the Molmo family, trained on the PixMo dataset of around one million carefully curated image–text pairs, using an OLMo-7B backbone with OpenAI CLIP as the vision encoder.
    \item \textbf{NVILA-Lite-2B}~\cite{liu2025nvila}: 
    NVILA-Lite-2B is a compact 2B-parameter visual language model built on the NVILA architecture, using a scale-then-compress design that processes high-resolution images and long videos efficiently by compressing visual tokens for faster inference and lower compute cost while maintaining competitive accuracy on standard benchmarks.
    \item \textbf{Qwen2.5-VL-3B-Instruct}~\cite{Qwen2.5-VL}: 
    Qwen2.5-VL-3B-Instruct is a 3B-parameter multi-modal vision-language model from the Qwen2.5-VL family, designed to process images, documents, and videos together with text. It combines a Vision Transformer encoder with a Qwen2.5-series language decoder to support instruction-following tasks such as OCR, document understanding, and general visual reasoning.
    \item \textbf{RoboRefer-2B-SFT}~\cite{zhou2025roborefer}: 
    RoboRefer-2B-SFT is a 2B-parameter vision-language model for robotics that is supervised-finetuned on RefSpatial and related instruction-following and referring datasets to handle spatial referring instructions in complex 3D scenes. In the second SFT step, RefSpatial is reused with both RGB and RGB-D inputs so that the image encoder learns robust spatial understanding from RGB alone while using depth as an auxiliary training signal, enabling both RGB-only and RGB-D inference at test time.
    \item \textbf{Qwen3-VL-235B-A22B-Instruct}~\cite{bai2025qwen3}:
    Qwen3-VL-235B-A22B-Instruct is a large open-weight Mixture-of-Experts vision–language model (235B parameters, 22B active) that combines text generation with visual understanding over images and video. It is an instruction-tuned Qwen3-VL variant designed for general-purpose multimodal tasks such as visual question answering, document parsing, and multilingual OCR in chat-style interactions.
\end{itemize}

\subsection{Training Data Sources}
\label{app:training_data}
A number of benchmarks and datasets for spatial understanding in VLMs have been proposed by the community.
Rather than generating training data from scratch, we leverage existing datasets and compose data mixes at varying scales to train the models.
Below we describe each dataset used in our experiments.

\begin{itemize}
    \item \textbf{SAT}~\cite{ray2025sat}:
    SAT is a synthetic spatial reasoning dataset built in the ProcTHOR-10K~\cite{deitke2022proc} interactive 3D indoor simulation environment, using about 22K procedurally generated apartment scenes composed of ~1K object assets and rendered into 2D views. 
    It contains 175K automatically generated question–answer pairs over 20K scenes, constructed from perfect 3D geometry and simulator metadata without human annotation, and is split into 127K static spatial QAs (relative position, depth, counting) and 48K dynamic spatial QAs grouped into Egocentric Movement, Object Movement, Allocentric Perspective, Goal Aiming, and Action Consequence, where actions in the simulator change spatial relationships across frames.
    \item \textbf{RoboSpatial}~\cite{song2025robospatial}:
    RoboSpatial is a large-scale 2D/3D spatial reasoning dataset built from real indoor and tabletop environments, where egocentric RGB images are paired with 3D scans instead of a synthetic simulator. 
    The data are collected as 3D scene scans and corresponding first-person images, and then automatically annotated with around 3M spatial relations over 1M images and 5K scans, capturing rich object–object and object–space relationships relevant for robotics. 
    The benchmark defines tasks such as spatial affordance prediction (where an object can be placed or an action can be executed), spatial relationship prediction (e.g., left/right, in front/behind, on/under), and robot manipulation tasks that test whether models can use these spatial cues to guide real-world actions.
    \item \textbf{SPAR-7M}~\cite{zhang2025flatland}:
    SPAR-7M is a large-scale spatial reasoning dataset built from indoor 3D scenes (e.g., ScanNet~\cite{dai2017scannet}, ScanNet++~\cite{yeshwanth2023scannet++}, Structured3D~\cite{zheng2020structured3d}) with around 7M QA pairs and 33 tasks covering a wide range of spatial perception and reasoning skills. 
    For our experiments, we sample from the following tasks: (1) Multi-view object spatial relation, which requires describing object–camera spatial relations in a multi-view setting (2) Single-view object spatial relation, a single-view multiple-choice task that asks models to select the correct relative position between two objects (3) Single-view spatial imagination, which evaluates single-view spatial imagination by asking models to verbally infer observer-centric relations beyond the directly visible configuration (4) Object count, which focuses on numerical reasoning by predicting object counts in the scene and (5) Multi-view spatial imagination, a multi-view spatial imagination task that requires describing how object–object relations change as the camera moves.

    \item \textbf{RefSpatial}~\cite{zhou2025roborefer}: 
    RefSpatial is a large‑scale spatial referring dataset built from 2D web images (OpenImages~\cite{kuznetsova2020open}) and 3D embodied videos (CA‑1M~\cite{lazarow2025cubify}), plus simulated scenes from Infinigen~\cite{raistrick2024infinigen} with Objaverse~\cite{deitke2023objaverse} assets. 
    It contains 2.5M RGB‑D samples and 20M QA pairs over 31 spatial relations and up to 5 reasoning steps, covering qualitative/quantitative spatial QA and point‑based location/placement supervision. 
    The tasks include object location (pointing to a described object), free‑space placement (pointing to a feasible placement location), and multi‑step spatial reasoning with explicit intermediate steps on simulated scenes. 
    In our work, we sample from all RefSpatial sources but use only the RGB images and associated annotations, discarding depth maps.
    \item \textbf{PRISM}~\cite{deshpande2025graspmolmo}:
    PRISM is a large-scale synthetic task-oriented grasping dataset built in a procedurally generated tabletop simulation using ShapeNet-Sem~\cite{chang2015shapenet, savva2015semantically} objects, ACRONYM~\cite{eppner2021acronym} grasp annotations, and SceneSynthesizer-based scene composition~\cite{eppner2025scene_synthesizer}, where 2,300+ object instances are rendered in heavily randomized scenes with calibrated RGB-D views, natural language task instructions, and associated 6-DoF grasp poses. GPT-based pipelines generate and match grasp-centric descriptions and manipulation tasks to appropriate grasps, yielding hundreds of thousands of samples for training and evaluating language-conditioned grasp prediction on both seen and novel objects.    

\end{itemize}

\subsection{Data Mix Composition}
\label{app:data_mix}

We construct four training data mixes of increasing scale using the five spatial datasets described in Section~\ref{app:training_data}.
For the 80k through 800k mixes, each dataset contributes an equal number of samples.
For the 2M mix, the per-dataset allocation is adjusted to accommodate differences in total dataset size, with smaller datasets (\eg, SAT) included in full while larger ones (\eg, RefSpatial) are subsampled.
Table~\ref{tab:data_mix} summarizes the per-dataset sample counts at each scale.

\begin{table}[h]
    \centering
    \caption{\textbf{Per-dataset sample counts for each data mix scale.} The 80k--800k mixes use uniform allocation across datasets. The 2M mix uses all available samples from smaller datasets and subsamples larger ones (RefSpatial at ${\sim}3.3\%$, SAT and PRISM in full).}
    \label{tab:data_mix}
    \resizebox{\linewidth}{!}{
    \begin{tabular}{lrrrrr}
        \toprule
        \textbf{Dataset} & \textbf{Available} & \textbf{80k} & \textbf{400k} & \textbf{800k} & \textbf{2M} \\
        \midrule
        SAT~\cite{ray2025sat}                &    172,384 &  15,997 &  79,997 & 159,998 & 172,384 \\
        RoboSpatial~\cite{song2025robospatial} &    300,000 &  15,999 &  79,998 & 159,999 & 300,000 \\
        SPAR-7M~\cite{zhang2025flatland}     &    608,538 &  15,997 &  79,997 & 159,996 & 608,538 \\
        RefSpatial~\cite{zhou2025roborefer}  & 16,363,656 &  15,994 &  79,996 & 159,997 & 540,397 \\
        PRISM~\cite{deshpande2025graspmolmo} &    378,844 &  16,000 &  80,000 & 160,000 & 378,844 \\
        \midrule
        \textbf{Total} & & \textbf{79,987} & \textbf{399,988} & \textbf{799,990} & \textbf{2,000,163} \\
        \bottomrule
    \end{tabular}
    }
\end{table}

Within each dataset, samples are drawn proportionally across all constituent sub-files (\eg, the six task categories of SAT, the seven QA types of RefSpatial).
Detailed per-file sampling ratios and the ready-to-use data mix configurations will be publicly released.

\subsection{Benchmarks}
\label{app:benchmarks_details}
We evaluate on the following benchmarks which are designed to test VLM's spatial understanding ability.
\begin{itemize}
    \item \textbf{EmbSpatial-Bench}~\cite{du2024embspatial}:
    EmbSpatial-Bench is introduced to systematically evaluate and improve large vision-language models’ spatial understanding for embodied tasks, addressing the gap that most existing Visual Spatial Reasoning benchmarks are 2D, dataset-centric (e.g., COCO~\cite{lin2014microsoft}/VG~\cite{krishna2017visual}), and object-centric rather than agent-centric, thus misaligned with real navigation and manipulation settings. 
    To better reflect embodied scenarios, the authors construct EmbSpatial-Bench from 3D indoor environments in MP3D~\cite{chang2017matterport3d}, AI2-THOR~\cite{kolve2017ai2}, and ScanNet~\cite{dai2017scannet}, rendering egocentric RGB-D views and using camera parameters and 3D coordinates to automatically derive 2D bounding boxes and spatial relation triplets between objects. 
    They define six fundamental relations (above, below, left, right, close, far) in the agent’s egocentric coordinate system to cover all three axes, convert these relations into multiple-choice QA pairs, and apply automatic filtering based on bounding box statistics followed by human verification to ensure object recognizability, relation correctness, and plausibility of distractor options.
    \item \textbf{CV-Bench}~\cite{tong2024cambrian}:
    CV-Bench (CV-Bench-2D, CV-Bench-3D) is a vision-centric multiple-choice benchmark built by repurposing standard vision datasets ADE20K~\cite{zhou2019semantic}, COCO~\cite{lin2014microsoft}, and Omni3D~\cite{brazil2023omni3d} into VQA-style examples that probe fundamental 2D and 3D understanding. 
    All images come from these real-world datasets rather than a synthetic simulator, and each instance is manually inspected, resulting in 2,638 high-quality examples with natural-language questions and four-way answer choices.
    The 2D split focuses on classic perception skills such as spatial relationship reasoning and object counting, while the 3D split targets depth ordering and relative distance understanding derived from the rich 3D annotations of Omni3D.
    \item \textbf{BLINK}~\cite{fu2024blink}:
    BLINK is a benchmark that repurposes 14 classic computer vision datasets into 3,807 visually prompted multiple-choice questions to assess fine-grained visual perception abilities such as relative depth estimation, spatial reasoning, visual correspondence, forensics detection, and multi-view understanding in multimodal large language models. 
    The images span abstract diagrams, synthetic scenes, and real-world photographs, covering diverse settings from object-centric views to outdoor landscapes without relying on a single simulation engine, and each task is constructed by overlaying simple visual markers and natural-language questions on existing annotated datasets. 
    Among its tasks, the relative depth (Rel. Depth) and spatial relation (Spat. Rel.) settings require models to compare which of two or more marked points is closer or to reason about geometric and positional relations between regions in the image, providing a focused probe of low-level 3D and spatial understanding beyond object recognition.
\end{itemize}

\subsection{BLINK Confidence Intervals}
\label{app:blink_ci}
To contextualize performance differences on small BLINK subsets, we report Wilson 95\% confidence intervals for the two BLINK splits used in Table~\ref{tab:spatial_benchmarks}: Rel.\ Depth ($n{=}124$) and Spat.\ Rel.\ ($n{=}143$).

\newcommand{\wci}[2]{\,{\scriptsize[#1, #2]}}

\begin{table}[t!]
\centering
\caption{\textbf{BLINK performance with Wilson 95\% confidence intervals.} Point estimates (accuracy, \%) are shown with Wilson 95\% CIs for BLINK Rel.\ Depth ($n{=}124$) and Spat.\ Rel.\ ($n{=}143$). \textbf{Bold} marks the best point estimate within each column.}
\label{tab:blink_wilson_ci}
\small
\begin{tabular}{lcc}
\toprule
\textbf{Model} &
\textbf{Rel.\ Depth} ($n{=}124$) &
\textbf{Spat.\ Rel.} ($n{=}143$) \\
\midrule
Molmo-7B-O-0924 &
78.2~\wci{70.2}{84.6} &
70.6~\wci{62.7}{77.5} \\
~~~~+ 80k  &
72.6~\wci{64.1}{79.7} &
60.8~\wci{52.7}{68.5} \\
~~~~+ 400k &
72.6~\wci{64.1}{79.7} &
68.5~\wci{60.5}{75.6} \\
~~~~+ 800k &
75.0~\wci{66.7}{81.8} &
61.5~\wci{53.4}{69.1} \\
~~~~+ 2M   &
71.0~\wci{62.4}{78.2} &
69.2~\wci{61.2}{76.2} \\
\midrule
NVILA-Lite-2B &
64.5~\wci{55.8}{72.4} &
67.1~\wci{59.1}{74.3} \\
~~~~+ 80k  &
53.2~\wci{44.5}{61.8} &
74.1~\wci{66.4}{80.6} \\
~~~~+ 400k &
71.8~\wci{63.3}{78.9} &
63.6~\wci{55.5}{71.1} \\
~~~~+ 800k &
57.3~\wci{48.5}{65.6} &
65.0~\wci{56.9}{72.4} \\
~~~~+ 2M   &
70.2~\wci{61.6}{77.5} &
62.9~\wci{54.8}{70.4} \\
\midrule
RoboRefer-SFT-2B &
\textbf{84.7}~\wci{77.3}{90.0} &
79.7~\wci{72.4}{85.5} \\
\midrule
Qwen2.5-VL-3B &
68.6~\wci{59.9}{76.1} &
83.9~\wci{77.0}{89.0} \\
~~~~+ 80k  &
58.1~\wci{49.3}{66.4} &
79.7~\wci{72.4}{85.5} \\
~~~~+ 400k &
58.9~\wci{50.1}{67.1} &
78.3~\wci{70.9}{84.3} \\
~~~~+ 800k &
58.1~\wci{49.3}{66.4} &
79.0~\wci{71.6}{84.9} \\
~~~~+ 2M   &
53.2~\wci{44.5}{61.8} &
78.3~\wci{70.9}{84.3} \\
\midrule
Qwen3-VL-235B &
\textbf{84.7}~\wci{77.3}{90.0} &
\textbf{90.2}~\wci{84.2}{94.1} \\
\bottomrule
\end{tabular}
\vspace{-1.0em}
\end{table}

\phantomsection
\label{anchor:appC}
\section{Additional Details on \TUNNEL}
\label{app:spatialtunnel}

This section presents additional details for \TUNNEL, including scene setup, the VQA protocol, proprietary-model results, and the object-size variant.

\subsection{Scene Generation Details}
\label{app:blender}
All scenes in \TUNNEL are rendered using Blender.
\paragraph{\textbf{Object placement.}}
The tunnel has a square cross-section of $2\,\text{m} \times 2\,\text{m}$.
Each scene contains two objects placed at different depths, with $\text{obj}_1$ always farther from the camera than $\text{obj}_2$.
To vary image-plane layout while preserving ground-truth depth, each object is swept independently over 16 discrete angular positions on the tunnel cross-section.
Holding depth fixed while varying $\theta$ changes the object’s image-plane position without altering its distance from the camera.

This construction yields matched image pairs that differ only in 2D layout while preserving the underlying depth ordering.

\paragraph{\textbf{Randomization.}}
For each scene, we independently randomize the following factors:
\begin{itemize}
    \item \textbf{Shape.} Each object is instantiated as either a sphere or a cube.
    \item \textbf{Color.} Each object is assigned one of seven colors:
    red, green, blue, yellow, cyan, magenta, or black.
    Materials are implemented with a Principled BSDF shader. Surface roughness is sampled uniformly from $[0.05, 1.0]$. Objects are constrained to have distinct $(\text{color}, \text{shape})$ combinations.
    
    \item \textbf{Size.} In the phase-variation setting, the base sizes are $s_1=0.2$ for $\text{obj}_1$ and $s_2=0.1$ for $\text{obj}_2$, each multiplied by an independent random scale factor in $[1.0, 1.5]$. In the size-variation setting, sizes are controlled systematically as described in Section~\ref{app:objsize}.
    
    \item \textbf{Lighting.} We use a Nishita sky texture in Blender. The sun rotation is sampled uniformly from $[1.25\pi, 1.75\pi]$ radians, and the background intensity is fixed to $0.15$. 
\end{itemize}

\subsection{VQA Protocol}

Given a rendered image containing two objects ($\text{obj}_1$ and $\text{obj}_2$), the model is asked a binary depth-comparison question.
To control for wording effects and answer-polarity bias, we instantiate four question templates per image, varying both the queried object order and the direction of comparison:

\begin{enumerate}
    \item \textit{``Is the \{obj$_1$\} closer to the camera than the \{obj$_2$\}?''} \hfill GT: \textbf{No}
    \item \textit{``Is the \{obj$_2$\} closer to the camera than the \{obj$_1$\}?''} \hfill GT: \textbf{Yes}
    \item \textit{``Is the \{obj$_2$\} farther from the camera than the \{obj$_1$\}?''} \hfill GT: \textbf{No}
    \item \textit{``Is the \{obj$_1$\} farther from the camera than the \{obj$_2$\}?''} \hfill GT: \textbf{Yes}
\end{enumerate}

For each joint angular configuration $(\theta_1, \theta_2)$, we render 12 scene instances with independently randomized shape, color, size, and lighting, for a total of \(16 \times 16 \times 12 = 3{,}072\) images. With four question templates per image, this yields $12{,}288$ question-image pairs.  Unless otherwise noted, responses are evaluated using the probability-based protocol described in Section~\ref{sec:spatialtunnel_setup}. We then average the four template-level correctness scores to obtain a single score for each configuration cell.

\subsection{Proprietary Model Results}
\label{app:gpt}

We additionally evaluate three proprietary configurations on \TUNNEL: GPT-5.2~\cite{singh2025openaigpt5card} in its default configuration, GPT-5.2 with reasoning enabled, and Gemini-2.5-Pro~\cite{comanici2025gemini25pushingfrontier}.
Because token-level logits are not exposed by the Azure API endpoints we tested, we evaluate models using final \texttt{Yes}/\texttt{No} outputs and report exact-match accuracy. We instantiate four question templates per image and average the resulting accuracies.
Table~\ref{tab:proprietary} summarizes these proprietary results alongside the open-source baselines that are recomputed with exact-match accuracy.

Under the default setting, GPT-5.2 attains a mean exact-match accuracy of \(0.613\), with \(Acc_{\text{con}} = 0.673\) and \(Acc_{\text{ctr}} = 0.552\), yielding a gap of \(\Delta = 0.120\). The positive gap indicates better performance on perspective-consistent cells than on counter cells, matching the directional bias observed in the open-source models.

Enabling reasoning improves GPT-5.2 from \(0.613\) to \(0.953\) mean accuracy and reduces the gap from \(\Delta = 0.120\) to \(\Delta = 0.058\). 
Gemini-2.5-Pro also performs strongly, achieving \(0.919\) mean accuracy with a slightly negative gap, \(\Delta = -0.028\). Taken together, the proprietary-model results show that enabling reasoning in GPT-5.2 both improves exact-match accuracy and reduces the consistent-counter gap on \TUNNEL, while Gemini-2.5-Pro likewise exhibits a near-zero gap.

\begin{table}[t]
\centering
\caption{\textbf{\textit{Exact-match response accuracy} on \TUNNEL under final-output evaluation.}
All models in this table are scored using discrete \texttt{Yes}/\texttt{No} outputs and averaged over the four question templates.
These values are therefore not directly comparable to the logit-based correctness scores \(v\) reported in Section~\ref{sec:spatialtunnel}.}
\label{tab:proprietary}
\small
\begin{tabular}{lcccc}
\toprule
\textbf{Model} & $\mathbf{Acc_{\text{all}}}$ & $\mathbf{Acc_{\text{con}}}$ & $\mathbf{Acc_{\text{ctr}}}$ & $\mathbf{\Delta}$ \\
\midrule
\multicolumn{5}{l}{Molmo-7B} \\
~~~~base         & 0.581 & 0.718 & 0.434 & +0.284 \\
~~~~+ 80k        & 0.500 & 0.500 & 0.500 & +0.000 \\
~~~~+ 400k       & 0.502 & 0.693 & 0.308 & +0.384 \\
~~~~+ 800k       & 0.551 & 0.768 & 0.316 & +0.452 \\
~~~~+ 2M         & 0.731 & 0.795 & 0.673 & +0.121 \\
\midrule
\multicolumn{5}{l}{NVILA-Lite-2B} \\
~~~~base         & 0.468 & 0.510 & 0.419 & +0.091 \\
~~~~+ 80k        & 0.454 & 0.545 & 0.364 & +0.181 \\
~~~~+ 400k       & 0.794 & 0.971 & 0.607 & +0.364 \\
~~~~+ 800k       & 0.851 & 0.979 & 0.727 & +0.252 \\
~~~~+ 2M         & 0.977 & 0.998 & 0.955 & +0.043 \\
\midrule
RoboRefer-2B-SFT & 0.919 & 0.956 & 0.879 & +0.076 \\
\midrule
\multicolumn{5}{l}{Qwen2.5-VL-3B} \\
~~~~base         & 0.493 & 0.653 & 0.342 & +0.310 \\
~~~~+ 80k        & 0.495 & 0.610 & 0.388 & +0.222 \\
~~~~+ 400k       & 0.498 & 0.576 & 0.423 & +0.153 \\
~~~~+ 800k       & 0.504 & 0.606 & 0.406 & +0.200 \\
~~~~+ 2M         & 0.506 & 0.681 & 0.335 & +0.346 \\
\midrule
Qwen3-VL-235B    & 0.929 & 0.958 & 0.909 & +0.050 \\
\midrule
GPT-5.2 (default)    & 0.613 & 0.673 & 0.552 & +0.120 \\
GPT-5.2 (reasoning)  & 0.953 & 0.980 & 0.922 & +0.058 \\
Gemini-2.5-Pro   & 0.919 & 0.905 & 0.933 & $-$0.028 \\
\bottomrule
\end{tabular}
\end{table}

\subsection{Extending the Analysis to Object Size}
\label{app:objsize}

Section~\ref{sec:spatialtunnel} showed that many VLMs rely on vertical image position as a shortcut for depth.
We next examine whether the same failure mode extends to another cue, \textbf{\textit{object size}}.
In natural images, larger objects often appear closer to the camera.
If a model relies on this cue, its depth judgments should degrade when relative size conflicts with the true depth ordering.

To test this, we construct a size-controlled variant of \TUNNEL in which the two object sizes, denoted by \(s_1\) for \(\text{obj}_1\) and \(s_2\) for \(\text{obj}_2\), are anti-correlated under the constraint \(s_1 + s_2 = 0.4\).
Specifically, we sweep $s_1$ over 10 equal intervals, yielding 11 values in total, and set $s_2 = 0.4 - s_1$.
The object depths are held fixed throughout, with $\text{obj}_1$ always farther from the camera than $\text{obj}_2$.
As $s_1$ increases, the scene moves from a size-consistent regime, where the farther object is smaller, to a size-conflicting regime, where the farther object is larger than the nearer one.

\begin{figure}[t]
\centering
\includegraphics[width=\textwidth]{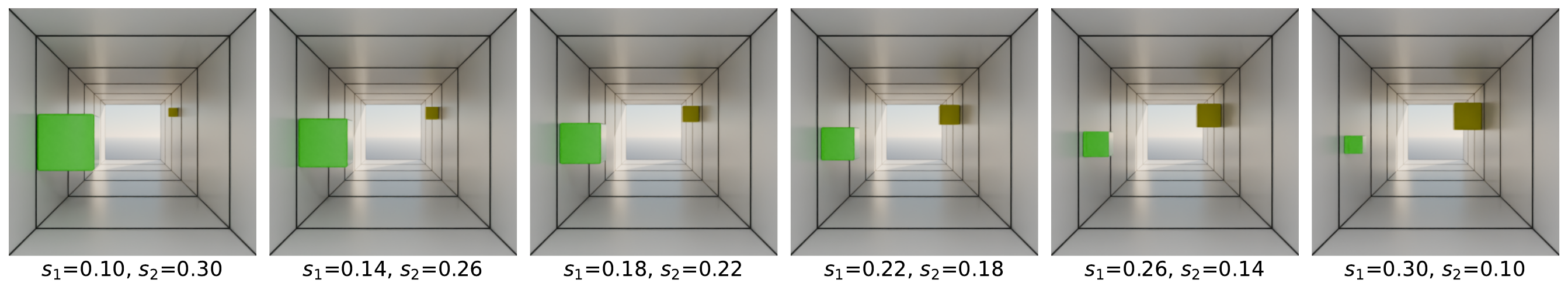}
\caption{\textbf{Object-size variation in \TUNNEL.} A representative scene rendered under six $(s_1, s_2)$ configurations with $s_1 + s_2 = 0.4$, where $s_1$ and $s_2$ denote the sizes of $\text{obj}_1$ and $\text{obj}_2$, respectively. $\text{obj}_1$ is always farther from the camera than $\text{obj}_2$. As $s_1$ increases from left to right, the farther object grows while the nearer object shrinks, moving from a size-consistent to a size-conflicting configuration. }
\label{fig:size_examples}
\end{figure}

Figure~\ref{fig:size_examples} shows a representative scene rendered under six $(s_1, s_2)$ configurations.
As $s_1$ increases from left to right, $\text{obj}_1$ grows while $\text{obj}_2$ shrinks correspondingly.
At the left endpoint ($s_1{=}0.10$, $s_2{=}0.30$), apparent size agrees with the true depth ordering.
At the right endpoint ($s_1{=}0.30$, $s_2{=}0.10$), the farther object appears substantially larger than the nearer one, which creates a strong cue that contradicts ground-truth depth.

We evaluate the same open-source VLMs as in Section~\ref{sec:finding} and report mean logit-based accuracy $v$ over all configurations.
Table~\ref{tab:size_variant} also reports accuracy at the two endpoints of the size sweep, $v_{s_1{=}0.1}$ and $v_{s_1{=}0.3}$.
We define the size-bias gap as
\[
\Delta_s = v_{s_1{=}0.1} - v_{s_1{=}0.3},
\]
which quantifies the \textit{size-distance entanglement}.
Larger positive values of $\Delta_s$ indicate stronger reliance on apparent size as a proxy for depth.

\begin{table}[t]
\centering
\caption{\textbf{Object-size variant results.} Mean logit-based accuracy $v$ across all $(s_1, s_2)$ configurations, accuracy at the two endpoints of the size sweep, and the size-bias gap $\Delta_s$. Larger positive $\Delta_s$ indicates stronger reliance on apparent size as a depth cue.}
\label{tab:size_variant}
\small
\begin{tabular}{lcccc}
\toprule
\textbf{Model} & $\mathbf{v}$ & $\mathbf{v_{s_1{=}0.1}}$ & $\mathbf{v_{s_1{=}0.3}}$ & $\mathbf{\Delta_s}$ \\
\midrule
\multicolumn{5}{l}{Qwen2.5-VL-3B} \\
~~~~base         & 0.510 & 0.507 & 0.515 & $-$0.008 \\
~~~~+ 80k        & 0.513 & 0.511 & 0.516 & $-$0.005 \\
~~~~+ 400k       & 0.500 & 0.494 & 0.510 & $-$0.016 \\
~~~~+ 800k       & 0.498 & 0.489 & 0.505 & $-$0.016 \\
~~~~+ 2M         & 0.522 & 0.528 & 0.509 & +0.018 \\
\midrule
\multicolumn{5}{l}{Molmo-7B} \\
~~~~base         & 0.589 & 0.609 & 0.553 & +0.057 \\
~~~~+ 80k        & 0.516 & 0.525 & 0.503 & +0.022 \\
~~~~+ 400k       & 0.605 & 0.656 & 0.532 & +0.124 \\
~~~~+ 800k       & 0.606 & 0.648 & 0.551 & +0.097 \\
~~~~+ 2M         & 0.801 & 0.888 & 0.643 & +0.246 \\
\midrule
\multicolumn{5}{l}{NVILA-Lite-2B} \\
~~~~base         & 0.515 & 0.530 & 0.487 & +0.043 \\
~~~~+ 80k        & 0.526 & 0.555 & 0.482 & +0.073 \\
~~~~+ 400k       & 0.727 & 0.815 & 0.586 & +0.229 \\
~~~~+ 800k       & 0.686 & 0.779 & 0.557 & +0.222 \\
~~~~+ 2M         & 0.828 & 0.895 & 0.689 & +0.207 \\
\midrule
RoboRefer-2B-SFT & 0.804 & 0.804 & 0.743 & +0.061 \\
\bottomrule
\end{tabular}
\end{table}

\begin{figure}[t]
\centering
\includegraphics[width=\textwidth]{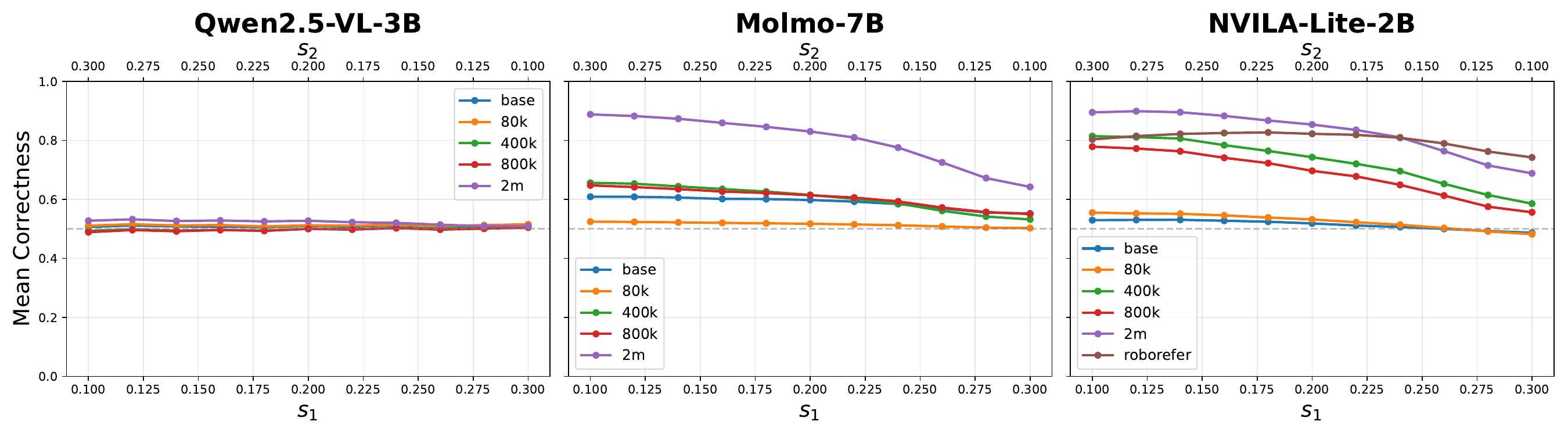}
\caption{\textbf{Correctness as a function of object size.} Mean logit-based correctness, averaged over all question templates, as a function of $\text{obj}_1$ size (bottom axis) and $\text{obj}_2$ size (top axis), with $s_1 + s_2 = 0.4$. Each curve corresponds to one training-data variant. Molmo and NVILA show clear degradation as the farther object becomes larger than the nearer one, whereas Qwen remains near chance throughout, indicating weak depth reasoning in this setting.}
\label{fig:size_correctness}
\end{figure}

Figure~\ref{fig:size_correctness} plots mean correctness as a function of $s_1$ (bottom axis) and $s_2 = 0.4 - s_1$ (top axis) for each model family.
Models that rely on the size cue show clear performance degradation as $s_1$ increases, that is, as the farther object becomes larger than the nearer one.
This confirms that apparent size acts as a confounding cue for depth, just as vertical position does in the main text.

\paragraph{\textbf{Analysis.}}
The object-size results closely mirror the vertical-distance entanglement results in the main text.
In both settings, performance is higher when a simple cue agrees with the true depth ordering, and lower when that cue is put in conflict with depth.

\begin{enumerate}
    \item \textbf{Qwen is largely insensitive to size, but remains near chance.}
    All Qwen variants cluster around chance performance ($v \approx 0.50$) and exhibit negligible gaps ($|\Delta_s| < 0.02$).
    This weak sensitivity should not be interpreted as robustness.
    Rather, it reflects limited depth discrimination in this setting.

    \item \textbf{Fine-tuning improves mean accuracy but often amplifies size-based shortcut reliance in Molmo and NVILA.}
    For Molmo-7B (2M), mean accuracy rises to $v = 0.801$, but the size-bias gap also grows to $\Delta_s = +0.246$.
    Similarly, NVILA-Lite-2B (2M) reaches $v = 0.828$ with $\Delta_s = +0.207$.
    This is the same qualitative pattern observed for vertical position: fine-tuning improves aggregate performance, yet also increases sensitivity to a correlated cue.

    \item \textbf{RoboRefer mitigates size bias while retaining high accuracy.}
    RoboRefer achieves $v = 0.804$, comparable to the strongest fine-tuned checkpoints, while exhibiting a much smaller gap ($\Delta_s = +0.061$). This mirrors the trend in the vertical-distance analysis, where RoboRefer also shows a relatively small gap, suggesting greater robustness to shortcut depth cues.
\end{enumerate}

Taken together with the vertical-position intervention, the size-variation results show the same qualitative pattern: accuracy is higher when a simple cue agrees with the true depth ordering, and lower when that cue conflicts with depth.
For the models we study, depth judgments therefore remain sensitive to multiple cues that often correlate with depth, including image height and apparent size.
These cues can support performance when they are aligned with the underlying 3D layout, but accuracy may decline when that correlation is weakened or reversed.
Thus, high average accuracy on depth queries should be interpreted with some caution, as it may not always reflect equally robust 3D spatial reasoning.

\phantomsection
\label{anchor:appD}
\section{Additional Details on Contrastive Probing}
\label{app:contrastive_probing_details}

In this section, we detail the swap pair construction methodology for each spatial category, describe the layer selection methodology, examine the cross-domain consistency of distance coherence, and present full heatmap and PCA visualizations for all model families.

\subsection{Swap Pair Construction}
\label{app:swap_pair_construct}
For horizontal and vertical pairs, we construct minimal contrastive pairs by symmetrically swapping the two queried objects: the question ``Is A to the left or right of B?'' becomes ``Is B to the left or right of A?'', which flips the ground-truth label while keeping all other visual context fixed.

For distance (far/close) pairs, the construction differs due to the format of EmbSpatial-Bench depth questions, which are presented as four-choice questions rather than direct relational queries. 
We identify the target object from the correct answer option, and sample the reference object uniformly at random from the remaining distractor options. 
The original question asks whether the target is far or close relative to the reference; the swapped question reverses these roles. 
This yields the same label-flip structure as the horizontal and vertical cases, while adapting to the available annotation format.

\hypertarget{appD_vdei}{}%
\subsection{A Brief Illustration of \VDEI}
\label{app:vd_ei_example}


VD-EI is positive when perspective-aligned pairs (above$\leftrightarrow$far, below$\leftrightarrow$close) are more similar than perspective-opposing pairs, and is near zero when the aligned and opposing terms largely cancel. In the extreme, VD-EI tends to be largest when aligned cosines are high while opposing cosines are negative (anti-aligned).

\hypertarget{app_layer}{}%
\subsection{Layer Selection Methodology}
\label{app:layer_selection}
This section provides supplementary details on layer selection criteria and per-model justifications.
The probing code extracts all layers; the user should select the appropriate $L^*$ from the saved outputs.

\subsubsection{Selection criteria.}

For each model family, we select a single representative layer $L^*$ at which to compute all probing metrics reported in the main text.
The selection is guided by the following criteria, applied in order of priority (\ie, if criteria conflict, earlier criteria take precedence):

\begin{enumerate}
    \item \textbf{Axis coherence plateau.} Coherence across all three spatial axes (horizontal, vertical, distance) is at or near its peak, indicating that stable axis-level structure has formed in the representation space.
    \item \textbf{VD-EI stability.} The VD-Entanglement Index should be at a meaningful plateau rather than in a transient region, ensuring that the selected layer captures the entanglement phenomenon we aim to analyze.
    When criteria conflict (e.g., \VDEI oscillates), we prioritize criterion (1) and select a layer from the shared high-coherence region across all three axes.
    \item \textbf{Avoidance of final layers.} The selected layer should not fall in the last few layers of the model, as these tend to be optimized for next-token prediction rather than rich representational encoding.
\end{enumerate}

\subsubsection{Supporting evidence for intermediate-layer selection.}

The preference for intermediate layers over final layers is well supported by prior work across both language and vision-language models.

Spatial representations in LLMs have been shown to form in early layers and plateau around the model midpoint; for instance, layer 50 out of 80 (63\%) in Llama-2-70b~\cite{touvron2023llama} was identified as the primary analysis layer for spatial probing~\cite{gurnee2024language}.
A systematic study across 32 probing tasks further confirms that intermediate layers encode richer representations than final layers, which become increasingly specialized for output generation~\cite{skean2025layer}.
In the multi-modal setting, visual layer selection experiments demonstrate that middle CLIP-ViT ~\cite{radford2021learning} layers outperform deep layers for spatial reasoning; on CV-Bench~\cite{tong2024cambrian}, layer 18 out of 24 outperforms the penultimate layer by 3\%, suggesting that vision-centric tasks such as spatial and positional reasoning benefit from mid-depth features rather than the deepest ones~\cite{chen2025rethinking}.

\subsubsection{Per-model selection.}
We describe the layer selection rationale for each model family below, with specific coherence and VD-EI values drawn from the full layer-wise trajectories.

\paragraph{\textbf{Molmo-7B-O-0924 (32 layers): $L^* = 23$ (72\%).}}
Coherence across all three axes peaks in the L20--25 range, with horizontal ${\sim}0.24$, vertical ${\sim}0.55$, and distance ${\sim}0.10$ at L23.
VD-EI reaches a plateau between L15 and L23 (0.5--0.7 for fine-tuned variants, 0.25 for base model), with L23 at the upper end of this range.
PCA visualizations confirm clearer cluster separation at L23 compared to neighboring layers.

\paragraph{\textbf{NVILA-Lite-2B (28 layers): $L^* = 20$ (71\%).}}
Vertical coherence plateaus between L18 and L25 (0.40--0.60 for fine-tuned variants, 0.80 for RoboRefer).
Horizontal coherence is stable across L15--27 (0.20--0.30 for fine-tuned variants, 0.65 for RoboRefer).
Distance coherence peaks between L18 and L25 (0.03--0.05 for fine-tuned variants, 0.18 for RoboRefer).
VD-EI plateaus at L17--26 (0.5--0.6 for fine-tuned variants), while RoboRefer shows around 0.25.
PCA visualizations are shown at $L{=}25$, where cluster separation is more visually distinguishable.

\paragraph{\textbf{Qwen2.5-VL-3B-Instruct (36 layers): $L^* = 28$ (78\%).}}
Horizontal coherence reaches ${\sim}0.37$ at L28.
Vertical coherence is ${\sim}0.35$ at L28, with a peak of 0.55 at L33.
Distance coherence remains low at ${\sim}0.035$.
VD-EI peaks at L20--22 (0.50), dips around L25, and rebounds to 0.50 at L28.
L28 balances meaningful entanglement with reasonable coherence while remaining below the output-specialized final layers.

\paragraph{\textbf{Qwen3-VL-235B-A22B-Instruct (94 layers): $L^* = 87$ (93\%).}}
Coherence across all axes forms very late in this model.
Horizontal coherence peaks at L83--87 (0.65), vertical peaks at L87--90 (0.63), and distance peaks at L85--90 (0.23).
VD-EI oscillates between 0.3 and 0.7 in the upper layers without a clear plateau.
The selected depth of 93\% notably exceeds the 71--75\% range observed in smaller models, likely reflecting architectural differences: Qwen3-VL-235B-A22B-Instruct is a 94-layer Mixture-of-Experts model with 235B total parameters (22B active), which may delay the formation of stable spatial representations to later layers.
We select $L^*=87$ from the shared high-coherence region across all three axes, despite \VDEI oscillation in the upper layers.
Importantly, the resulting cross-model $\CohD$ ranking is robust to alternative valid layer choices (see below).

\hypertarget{app_layer_robust}{}%
\subsection{Robustness to Alternative Layer Choices.}
Our layer selection follows the predefined protocol above and is applied independently per model family before cross-model comparison.
Candidate ranges are defined as the union of layers where $\CohH$, $\CohV$, and $\CohD$ are near peak; when the criteria differ, we prioritize high axis coherence.
To quantify sensitivity, we sample $1\mathrm{K}$ random layers within each candidate range (without refitting) and recompute the cross-model $\CohD$ ranking; the resulting rankings show high agreement with the reported ordering (Spearman $\rho=0.928$).


\hypertarget{app_cross}{}%
\subsection{Cross-Domain Consistency of Distance Coherence}
\label{app:cross_domain_distance_coherence}

\begin{figure}[t]
    \centering
    \includegraphics[width=\linewidth]{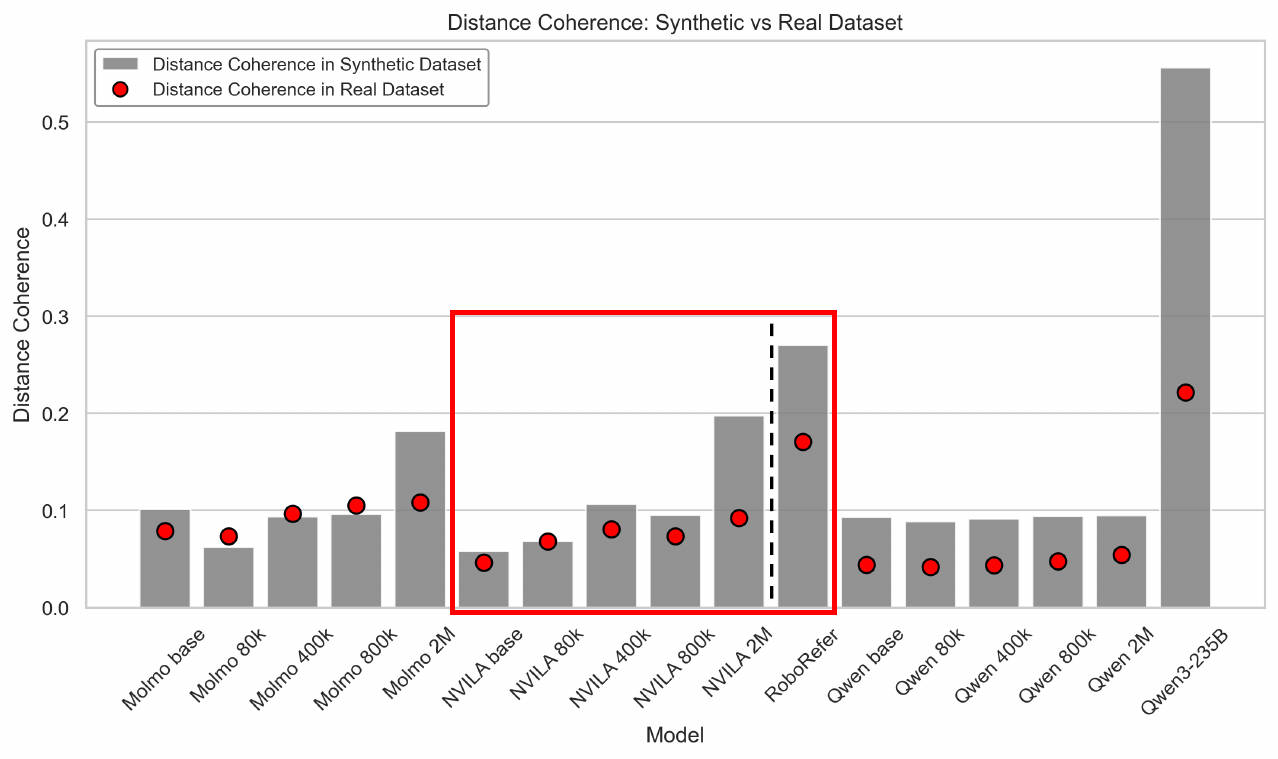}
    \caption{\textbf{Distance Coherence measured on synthetic (\TUNNEL) vs.\ real (EmbSpatial-Bench) datasets.} Gray bars denote $\mathrm{Coh}_D$ computed on \TUNNEL; red dots denote $\mathrm{Coh}_D$ on EmbSpatial-Bench. Although the absolute magnitudes differ across domains, the relative ordering within each model family is largely preserved. The red box highlights the NVILA family, where the ranking is identical across both datasets.}
    \label{fig:coherence_cross_domain}
\end{figure}

Figure~\ref{fig:coherence_cross_domain} compares $\mathrm{Coh}_D$ computed on the synthetic dataset (\TUNNEL) with that computed on EmbSpatial-Bench.
The absolute values differ between the two domains, as $\mathrm{Coh}_D$ in \TUNNEL is generally higher; however, the \emph{relative} ordering of models within each family is largely consistent.

For the NVILA family (highlighted in Figure~\ref{fig:coherence_cross_domain}), the ranking is preserved across both datasets:
RoboRefer ${>}$ NVILA\,2M ${>}$ NVILA\,400k ${\approx}$ NVILA\,800k ${>}$ NVILA\,80k ${>}$ NVILA\,base.
The Molmo family exhibits minor rank swaps among adjacent checkpoints, but the overall trend of increasing coherence with training scale, from 80k to 2M, is shared across domains.
For the Qwen family, the Qwen2.5-VL-3B scale variants cluster tightly in both settings, while Qwen3-235B shows a markedly different profile.

These results suggest that the absolute magnitude of $\mathrm{Coh}_D$ can be influenced by the environment, but it provides a \emph{reliable relative comparison} when models are evaluated under the same data condition.
We publicly release both evaluation datasets and the probing pipeline so that future users can benchmark new models under identical conditions and compare against the values reported in this paper, enabling $\mathrm{Coh}_D$ to serve as a reproducible measure of spatial representation quality.

\subsection{Heatmap and PCA Results}

This section presents the cross-category similarity heatmaps and PCA visualizations for all model families.
Similarity is computed as the cosine similarity between each category's mean delta vector.

\begin{figure}[t]
    \centering
    \includegraphics[width=\linewidth]{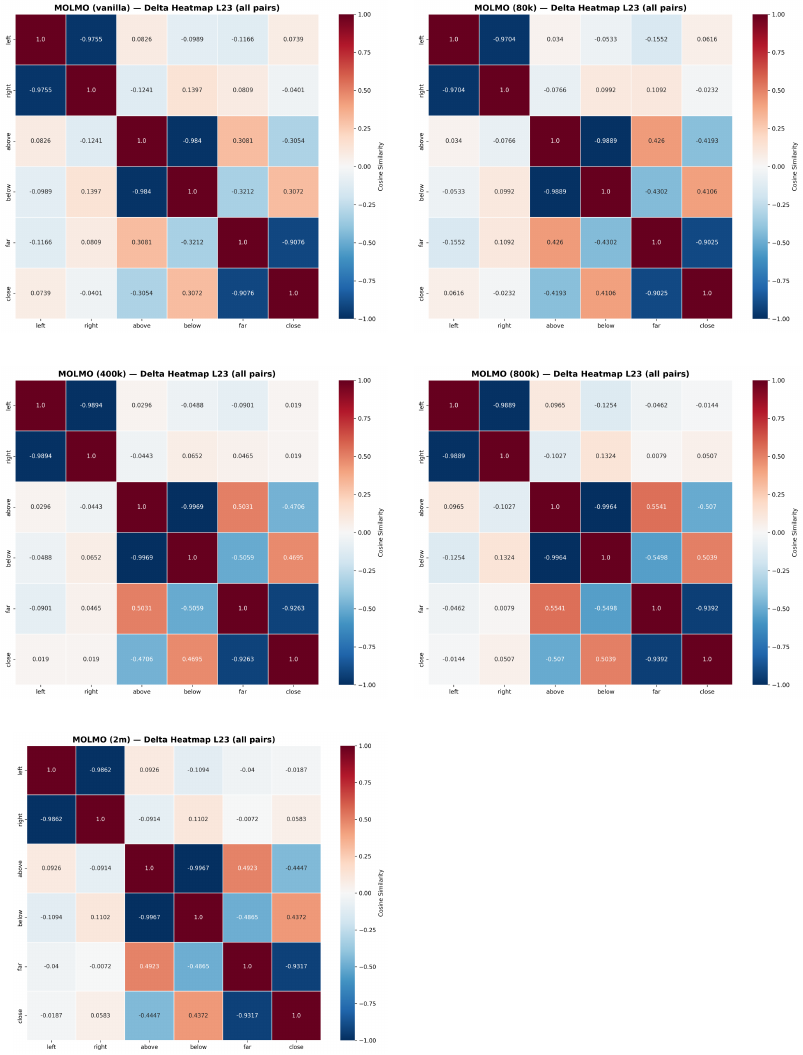}
    \caption{\textbf{Cross-category similarity heatmaps for the Molmo family.} Each cell shows the cosine similarity between mean delta vectors of two categories. Variants range from vanilla (base Molmo-7B) to 2M (SFT with 2M-sample data mix).}
    \label{fig:heatmap_molmo}
\end{figure}

\begin{figure}[t]
    \centering
    \includegraphics[width=\linewidth]{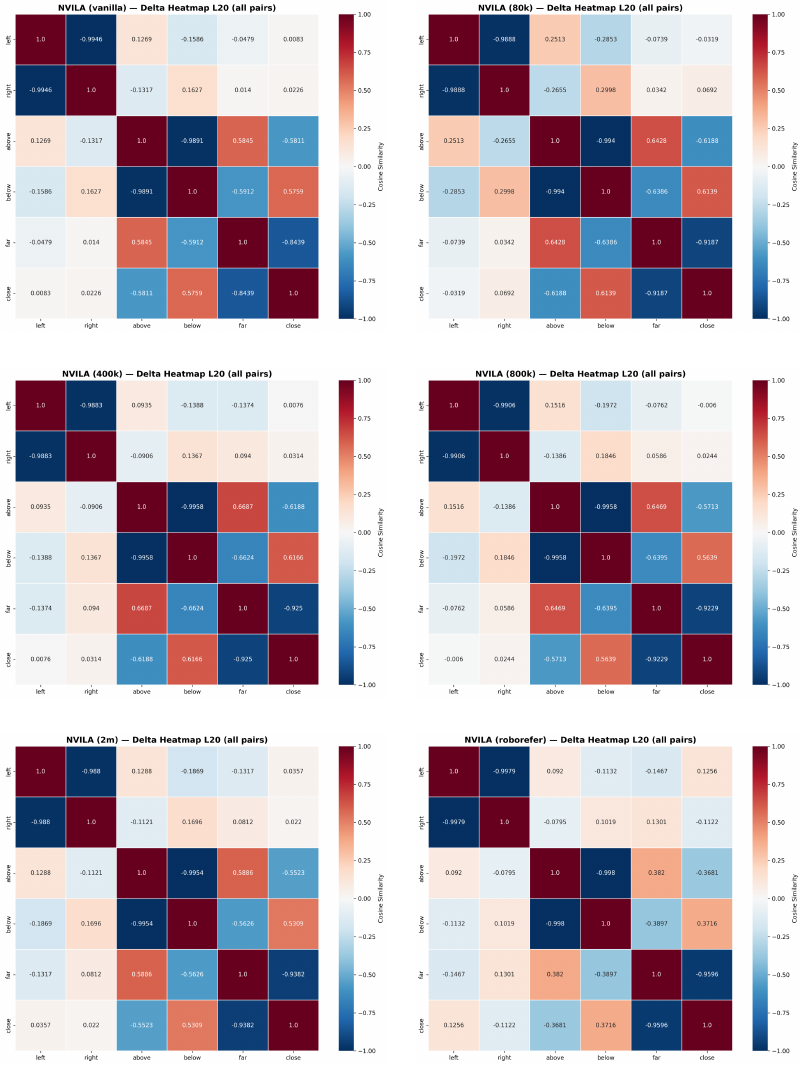}
    \caption{\textbf{Cross-category similarity heatmaps for the NVILA family.} Variants include NVILA-Lite-2B from vanilla (base) through 2M (SFT), plus RoboRefer (RoboRefer-2B-SFT).}
    \label{fig:heatmap_nvila}
\end{figure}

\begin{figure}[t]
    \centering
    \includegraphics[width=\linewidth]{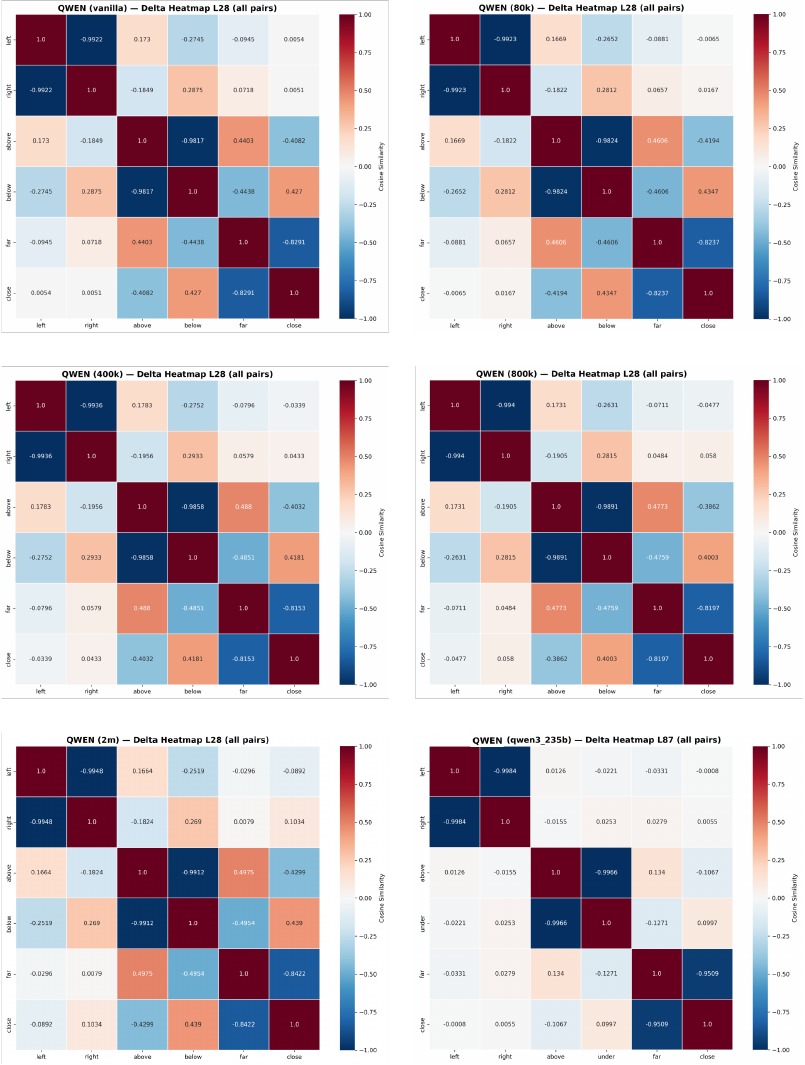}
    \caption{\textbf{Cross-category similarity heatmaps for the Qwen family.} Variants include Qwen2.5-VL-3B-Instruct (vanilla through 2M) and Qwen3-VL-235B-A22B-Instruct.}
    \label{fig:heatmap_qwen}
\end{figure}

As shown in the heatmap results (Figure~\ref{fig:heatmap_molmo}, \ref{fig:heatmap_nvila}, \ref{fig:heatmap_qwen}), the similarity between opposing categories on the same axis (\eg, \emph{left} and \emph{right} on the horizontal axis) is consistently near $-1$, indicating that the model encodes opposite spatial directions as antiparallel vectors in representation space.
Additionally, similarity between horizontal and the other axes (\ie, vertical and distance) is close to zero, suggesting that the horizontal axis is encoded independently from both vertical and distance representations.
However, between the vertical and distance axes, the perspective-aligned pairs \emph{above}$\leftrightarrow$\emph{far} and \emph{below}$\leftrightarrow$\emph{close} exhibit meaningful positive similarity in the range of 0.1--0.65 across all models.
This confirms that vertical and distance representations are directionally coupled, consistent with the entanglement phenomenon described in the main text.

\begin{figure}[t]
    \centering
    \includegraphics[width=\linewidth]{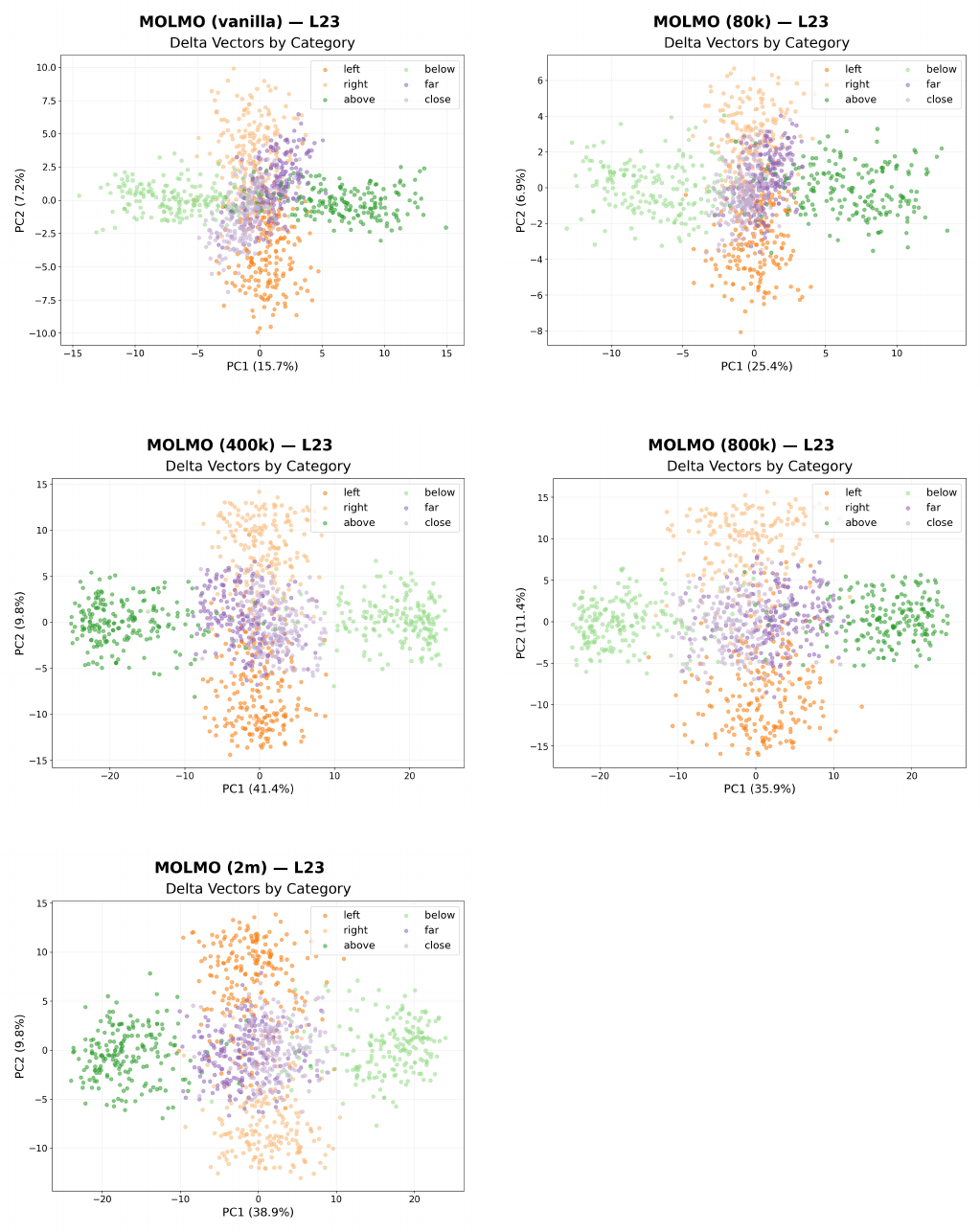}
    \caption{\textbf{2D PCA of delta vectors for the Molmo family.} Each point represents a per-sample delta vector, colored by spatial category. Opposing categories (\eg, \emph{left} vs.\ \emph{right}) separate along shared principal components, while \emph{far}/\emph{close} overlap with \emph{above}/\emph{below}, reflecting vertical-distance entanglement.}
    \label{fig:pca_2d_molmo}
\end{figure}
\begin{figure}[t]
    \centering
    \includegraphics[width=\linewidth]{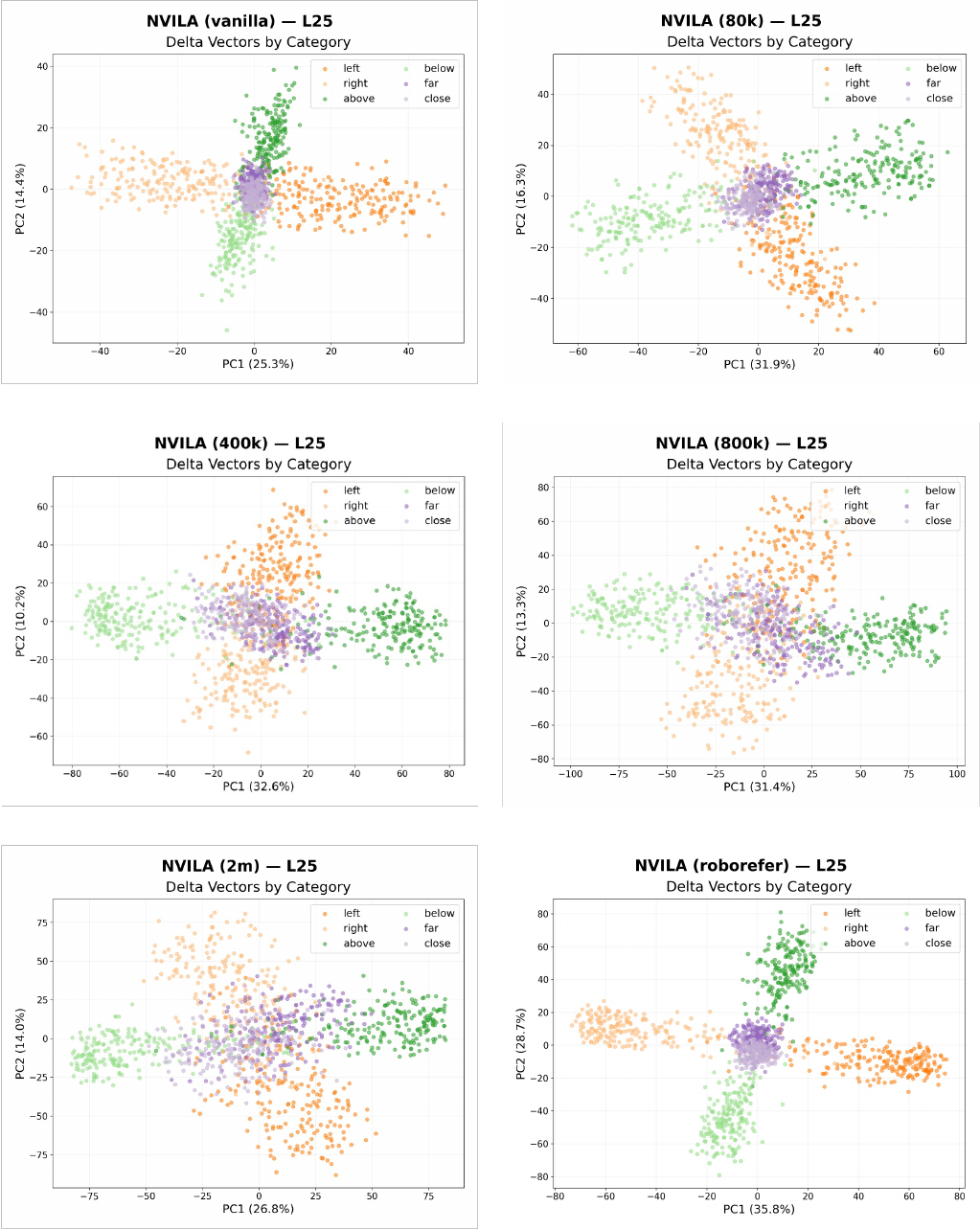}
    \caption{\textbf{2D PCA of delta vectors for the NVILA family.} RoboRefer shows notably tighter distance-axis clusters (\emph{far}/\emph{close}) separated from vertical categories, consistent with its higher $\mathrm{Coh}_D$ and lower VD-EI.}
    \label{fig:pca_2d_nvila}
\end{figure}
\begin{figure}[t]
    \centering
    \includegraphics[width=\linewidth]{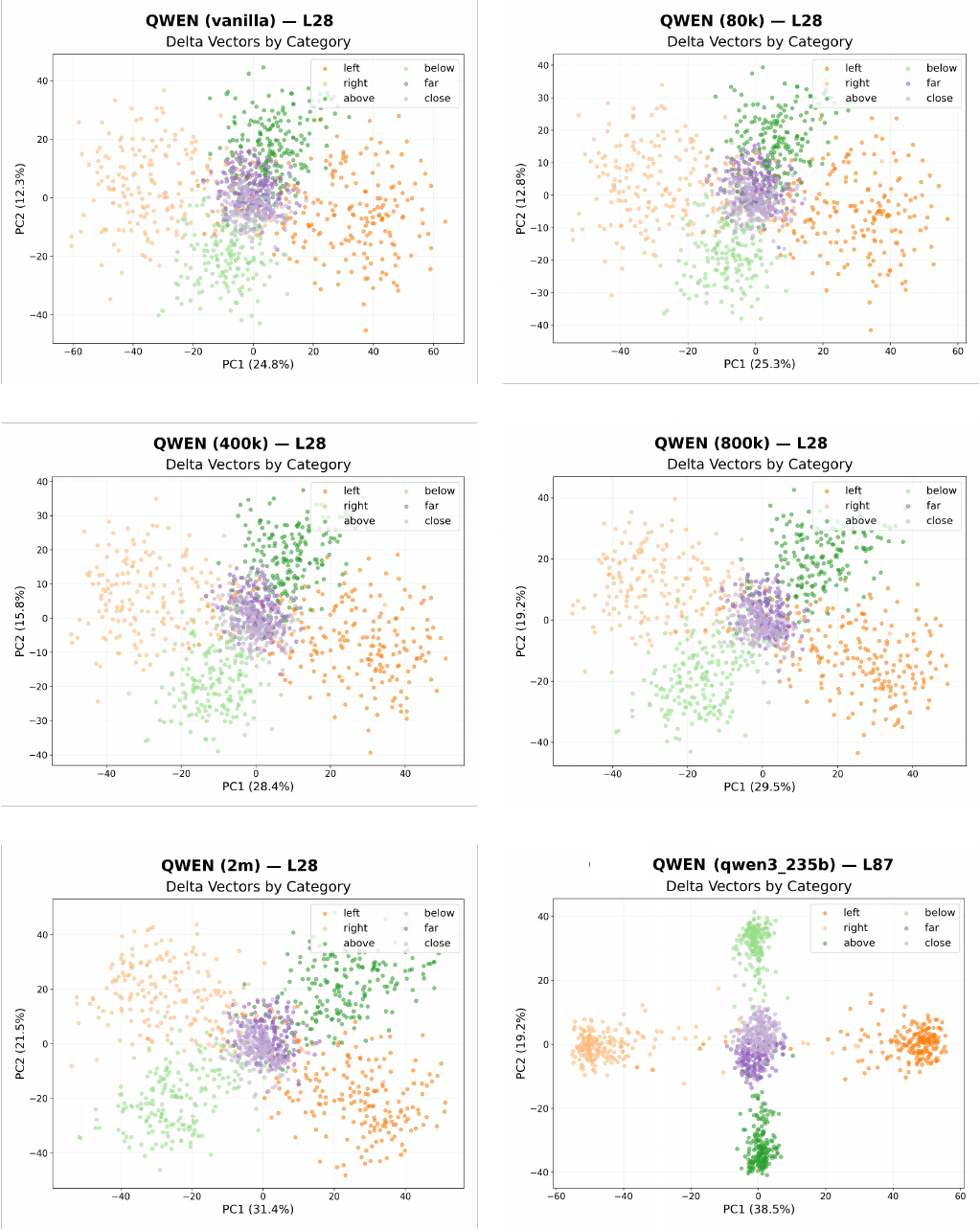}
    \caption{\textbf{2D PCA of delta vectors for the Qwen family.} Variants include Qwen2.5-VL-3B-Instruct and Qwen3-VL-235B-A22B-Instruct. Qwen3-VL-235B exhibits markedly cleaner cluster separation across all three axes.}
    \label{fig:pca_2d_qwen}
\end{figure}
\begin{figure}[t]
    \centering
    \includegraphics[width=0.8\linewidth]{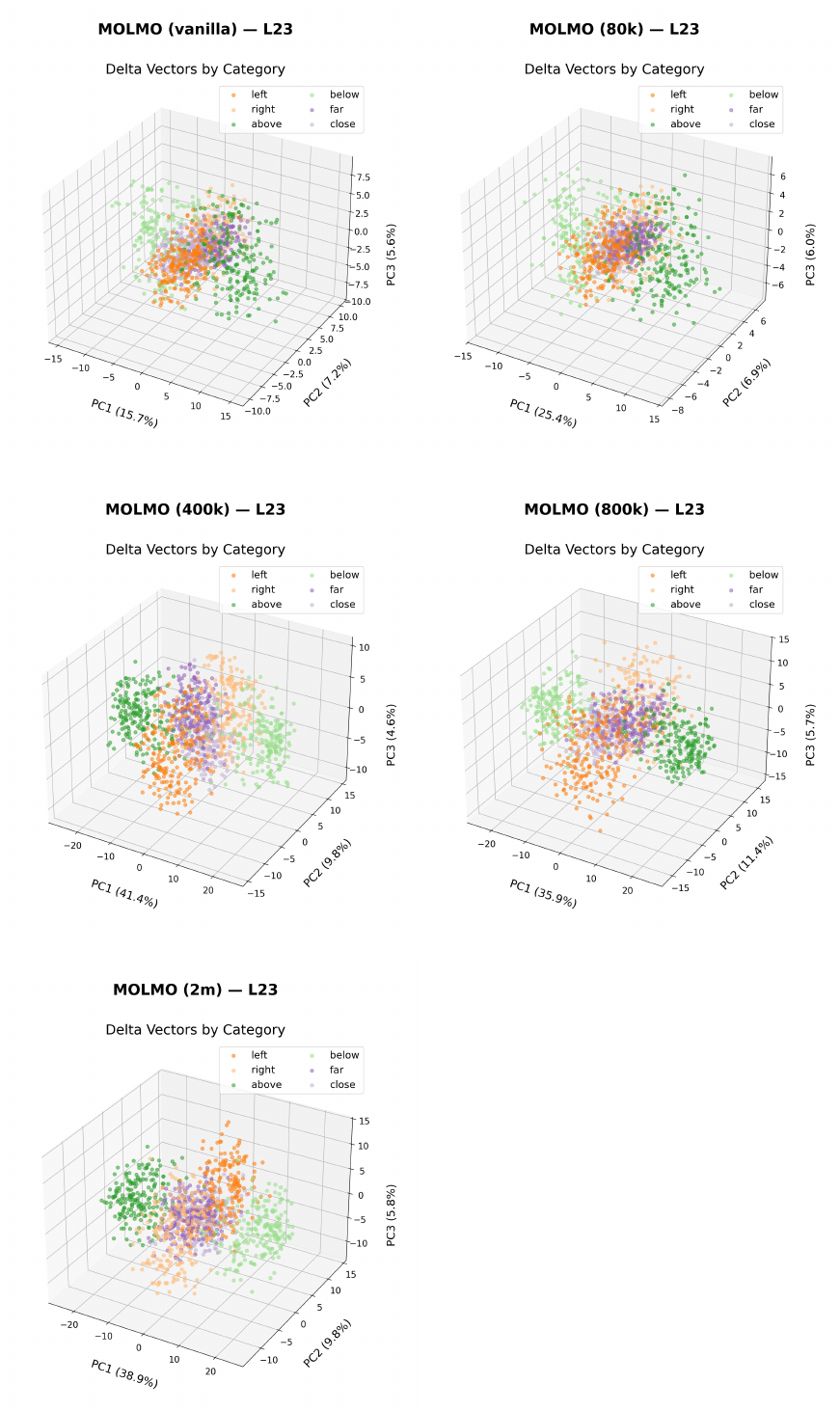}
    \caption{\textbf{3D PCA of delta vectors for the Molmo family.} A distinct distance axis does not clearly emerge, although delta vectors in the horizontal and vertical axes appear more well-clustered with data scaling.}
    \label{fig:pca_3d_molmo}
\end{figure}
\begin{figure}[t]
    \centering
    \includegraphics[width=0.8\linewidth]{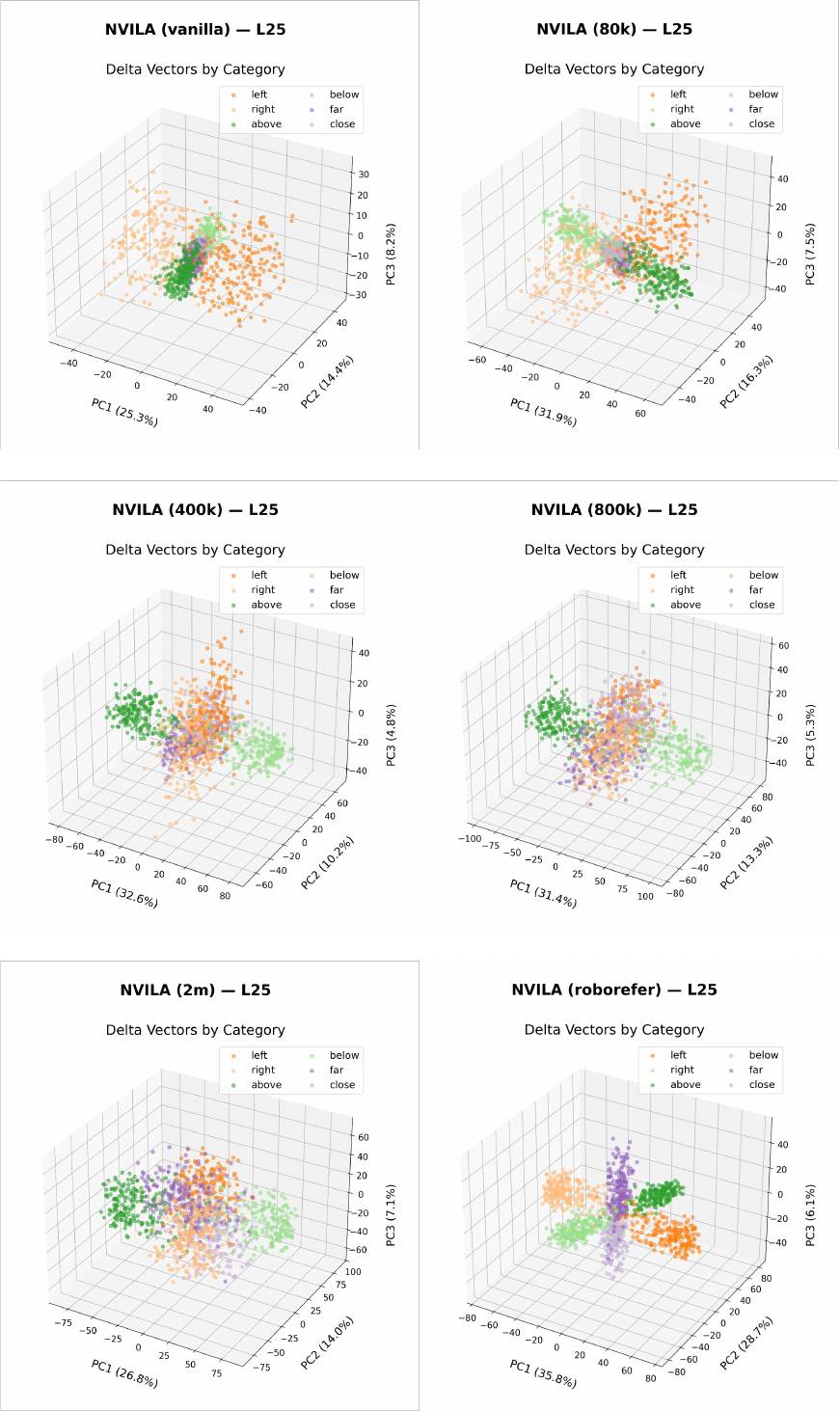}
    \caption{\textbf{3D PCA of delta vectors for the NVILA family.} RoboRefer's distance clusters (\emph{far}/\emph{close}) occupy a distinct subspace from vertical categories, unlike the fine-tuned variants.}
    \label{fig:pca_3d_nvila}
\end{figure}
\begin{figure}[t]
    \centering
    \includegraphics[width=0.8\linewidth]{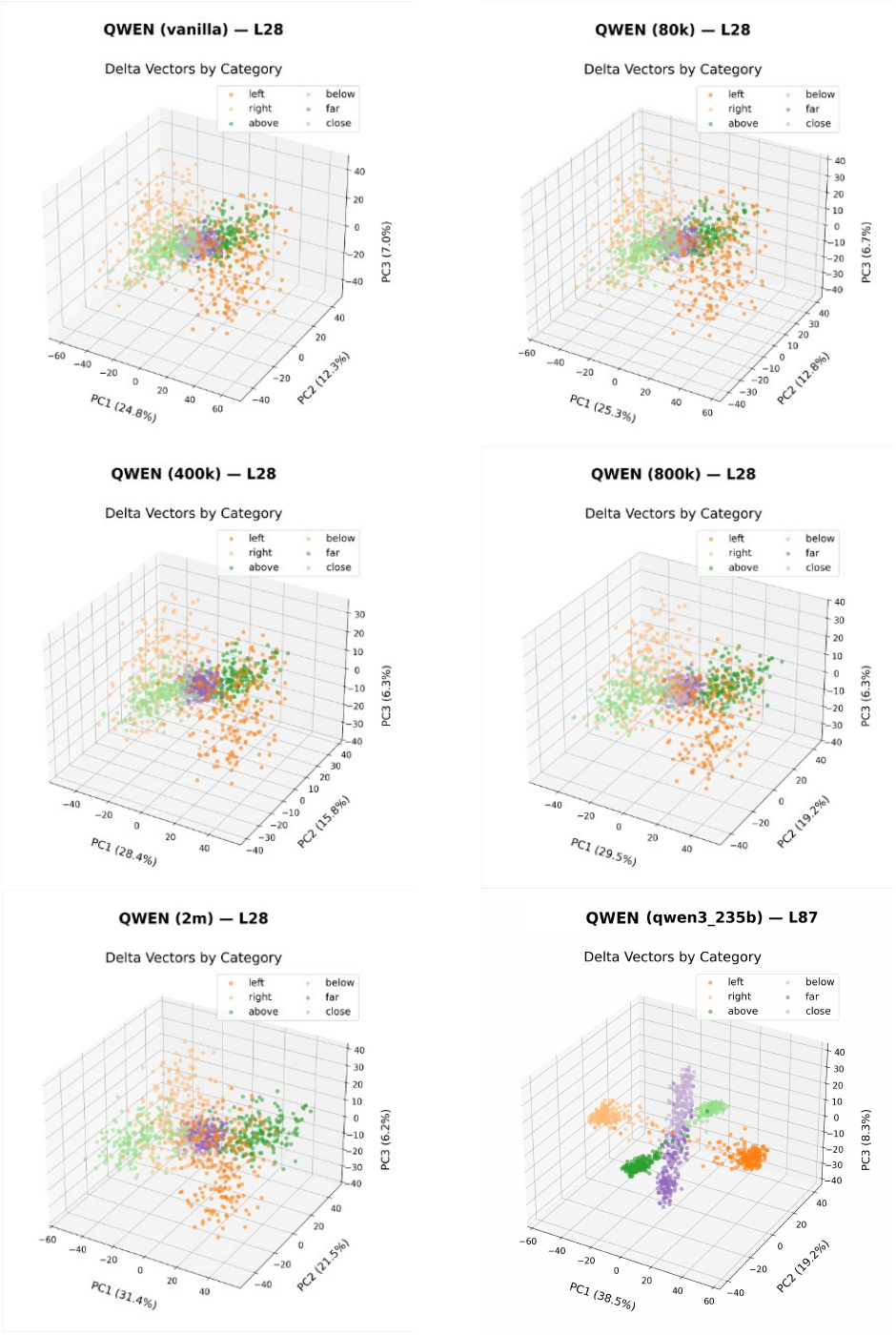}
    \caption{\textbf{3D PCA of delta vectors for the Qwen family.} Variants include Qwen2.5-VL-3B-Instruct and Qwen3-VL-235B-A22B-Instruct. Qwen3-VL-235B shows clear three-way separation among horizontal, vertical, and distance axes in 3D space.}
    \label{fig:pca_3d_qwen}
\end{figure}

\end{document}